\documentclass[10pt,twocolumn,letterpaper]{article}

\usepackage{iccv}
\usepackage{times}
\usepackage{epsfig}
\usepackage{graphicx}
\usepackage{amsmath}
\usepackage{amssymb}
\usepackage{multirow}
\usepackage{booktabs}
\usepackage{bbding}
\usepackage{amsthm,amsfonts,bm,bbm}
\usepackage{arydshln}
\usepackage{algorithm}
\usepackage{listings}
\usepackage{dsfont}
\usepackage{subfig}
\usepackage{colortbl}
\definecolor{mygray}{gray}{.9}
\usepackage[accsupp]{axessibility}  % Improves PDF readability for those with disabilities.

\usepackage{color}
\usepackage{etoolbox}
\makeatletter
\AfterEndEnvironment{algorithm}{\let\@algcomment\relax}
\AtEndEnvironment{algorithm}{\kern2pt\hrule\relax\vskip3pt\@algcomment}
\let\@algcomment\relax
\newcommand\algcomment[1]{\def\@algcomment{\footnotesize#1}}
\renewcommand\fs@ruled{\def\@fs@cfont{\bfseries}\let\@fs@capt\floatc@ruled
  \def\@fs@pre{\hrule height.8pt depth0pt \kern2pt}%
  \def\@fs@post{}%
  \def\@fs@mid{\kern2pt\hrule\kern2pt}%
  \let\@fs@iftopcapt\iftrue}
\makeatother
\usepackage[pagebackref=true,breaklinks=true,letterpaper=true,colorlinks,bookmarks=false]{hyperref}

\iccvfinalcopy % *** Uncomment this line for the final submission

\def\iccvPaperID{4989} % *** Enter the ICCV Paper ID here

\ificcvfinal\fi
\newcommand{\modelname}{STA\xspace}

\begin{document}

%%%%%%%%% TITLE
\title{Prune Spatio-temporal Tokens by Semantic-aware Temporal Accumulation}

\author {
    % Authors
    Shuangrui Ding\textsuperscript{\rm 1,3}\thanks{Work done during an internship at Huawei Cloud.} \quad
    Peisen Zhao\textsuperscript{\rm 2} \quad Xiaopeng Zhang\textsuperscript{\rm 2}  \quad Rui Qian\textsuperscript{\rm 3} \quad 
    Hongkai Xiong\textsuperscript{\rm 1} \quad 
    Qi Tian\textsuperscript{\rm 2}\thanks{ Corresponding author. Email: tian.qi1@huawei.com.}\\
    \textsuperscript{\rm 1}Shanghai Jiao Tong University \quad 
    \textsuperscript{\rm 2}Huawei Cloud  \quad  \textsuperscript{\rm 3}The Chinese University of Hong Kong   \\
    {\tt\small \{dsr1212, xionghongkai\}@sjtu.edu.cn qr021@ie.cuhk.edu.hk}  \\  
    {\tt\small \{pszhao93, zxphistory\}@gmail.com  tian.qi1@huawei.com}
}

\maketitle
% Remove page # from the first page of camera-ready.
\ificcvfinal\thispagestyle{empty}\fi
% Vit-S: 4.9821 4.8932 4.6757
% Vit-B: 5.3922 5.3385 5.1345
% Vit-L: 4.0702 3.9404 3.5650
% Vit-H: 4.8542 4.7573 4.3700
%%%%%%%%% ABSTRACT
\begin{abstract}
Transformers have become the primary backbone of the computer vision community due to their impressive performance. However, the unfriendly computation cost impedes their potential in the video recognition domain. 
To optimize the speed-accuracy trade-off, we propose \textbf{S}emantic-aware \textbf{T}emporal \textbf{A}ccumulation score (STA) to prune spatio-temporal tokens integrally. 
STA score considers two critical factors: temporal redundancy and semantic importance. The former depicts a specific region based on whether it is a new occurrence or a seen entity by aggregating token-to-token similarity in consecutive frames while the latter evaluates each token based on its contribution to the overall prediction. 
As a result, tokens with higher scores of STA carry more temporal redundancy as well as lower semantics thus being pruned. 
Based on the STA score, we are able to progressively prune the tokens without introducing any additional parameters or requiring further re-training. We directly apply the STA module to off-the-shelf ViT and VideoSwin backbones, and the empirical results on Kinetics-400 and Something-Something V2 achieve over $30\%$ computation reduction with a negligible $\sim0.2\%$ accuracy drop. The code is released at \href{https://github.com/Mark12Ding/STA}{https://github.com/Mark12Ding/STA}.
\end{abstract}

%%%%%%%%% BODY TEXT
\section{Introduction}
% In the last decade, convolutional neural networks (ConvNets)~\cite{lecun1989backpropagation,krizhevsky2012imagenet,he2016deep,hu2018squeeze,radosavovic2020designing,liu2022convnet} play an important role as the \textit{de-facto} backbone architecture for general computer vision tasks. 
% ConvNets enjoy desirable properties like translation equivariance and locality due to the sliding window mechanism. 
% Those essential inductive biases make ConvNets widely adopted on image recognition~\cite{he2016deep,howard2017mobilenets}, action recognition~\cite{simonyan2014two,feichtenhofer2019slowfast}, semantic segmentation~\cite{long2015fully,chen2017deeplab}, and object detection~\cite{ren2015faster,lin2017focal}, and attain ideal performance in terms of efficiency and accuracy.  
Recently, there has been an unstoppable shift in the general backbone design from Convolutional Neural Networks (ConvNets) to Transformers, which are originally employed in natural language processing, and has shown promising potential for various vision tasks~\cite{dosovitskiy2021an, zhao2021point, zhang2021vidtr, liu2021swin, Arnab_2021_ICCV, tan2021relaxed, chen2023generative}. 
The key component of Transformers is the self-attention mechanism, which is apt to capture long-range dependencies and empowers ViT to perceive the global reception field. 
The seminal work, Vision Transformer (ViT)~\cite{dosovitskiy2021an} closely follows the original Transformer
architecture~\cite{vaswani2017attention}.
Equipped with a large-scale model and dataset, ViT outperforms ConvNets in image classification by a considerable margin. 
Inspired by this superior scaling behavior, Transformers have gained popularity as a backbone choice and are widely adopted for image recognition~\cite{liu2021swin, tu2022maxvit}, action recognition~\cite{Arnab_2021_ICCV, liu2022video}, semantic segmentation~\cite{zheng2021rethinking, cheng2021per}, action detection~\cite{wu2022memvit, Herzig_2022_CVPR}, temporal perception~\cite{tan2023temporal, tan2022pointtad}, \emph{etc.}
% , which directly grids the image into several patches and flattens the 2D patches into 1D-sequence. 
\begin{figure}
    \centering
    \includegraphics[width=\linewidth]{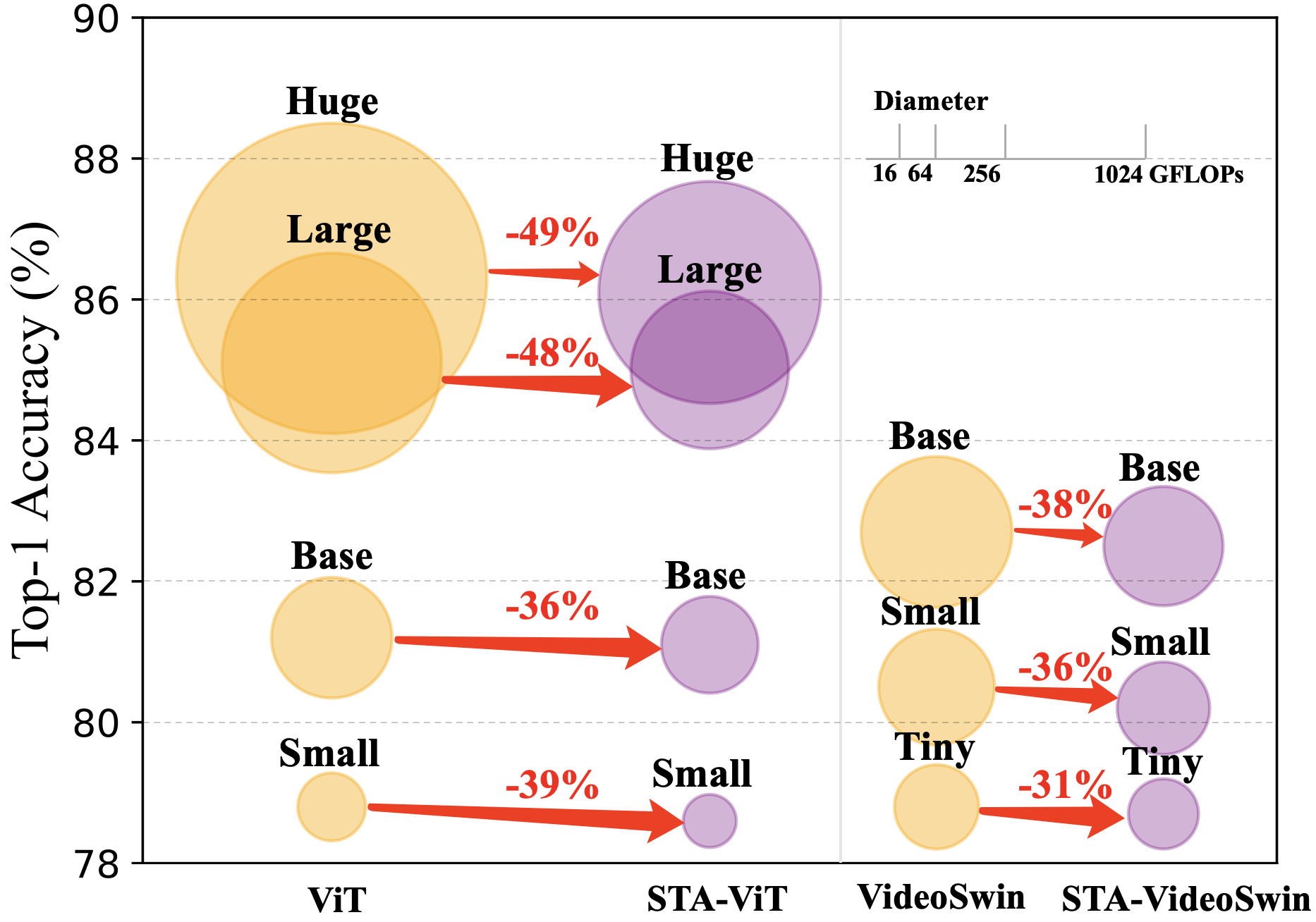}
    \caption{Kinectics-400 result for ViT and VideoSwin. The bubble’s area is proportional to FLOPs of a variant in a model family. ViT/VideoSwin here takes $16/32\times 224^2$ video. The proposed STA saves over $30\%$ FLOPs for all model variants with a negligible drop in performance.}
    % \PS{Why is the circle of Swin smaller than ViT, as Swin is 32 frames, and the FLOPs data in the experiment do not match?} \DSR{The circle of ViT-small is much smaller than Swin-small.}
    \label{fig:sota}
\end{figure}
   
Despite the promising potential of Transformers in spatio-temporal vision tasks, such as action recognition, the quadratic increase in complexity caused by the temporal dimension makes the video Transformers computationally unfriendly compared to images. 
For instance, earlier work TimeSformer~\cite{bertasius2021space}, which applies Transformer backbone for video, required $7.14$ Tera FLOPs to achieve $80.7\%$ accuracy on the Kinetics-400 action recognition benchmark. The excessive computational cost makes it impractical for deployment in real-world scenarios. Therefore, there is an urgent need to explore ways to profit from the performance gains of Transformers while maintaining an affordable computation burden.
% Nevertheless, the computation burden hinders the application of Transformers in many spatio-temporal vision tasks, \textit{e.g.,} action recognition.
% This issue originates from the fact that the temporal dimension linearly grows the complexity and usually incurs unfriendly computation compared to the image. 
% Earlier work TimeSformer~\cite{bertasius2021space}, that applies Transformer backbone for video, takes $7.14$ Tera FLOPs to obtain 80.7\% accuracy at Kinectics-400 action recognition benchmark~\cite{kay2017kinetics}. The giant amount of FlOPs makes it inaccessible to deploy for practical usage. Therefore, how to profit from the performance gain of transformers, while maintaining an affordable computation overhead becomes an increasingly urgent need. 
%motivates the video community to develop efficient video transformers. 

Fortunately, Transformers can handle a flexible number of tokens as input. Recent attempts~\cite{ryoo2021tokenlearner, rao2021dynamicvit, wang2021not}, that dynamically prune tokens for images, have remarkably reduced computation overhead. These pruning approaches have inspired us to explore token pruning in the video domain, so as to balance accuracy and computation costs accordingly. However, performing frame-wise pruning alone seems to be not optimal since it ignores temporal context and disrupts the dynamic structure of the video. To address this issue, recent work STTS~\cite{wang2022efficient} decouples token pruning into temporal and spatial dimensions. Specifically, STTS first drops meaningless frames and then filters out detail-rich regions from the remained frames. However, this spatio-temporal decoupling strategy lacks contextual modeling of continuous temporal information, leading to limited performance.

In this paper, we argue that pruning spatio-temporal tokens integrally can lead to further computation reduction at an acceptable cost of accuracy degradation. To this end, we propose the \textbf{S}emantic-aware \textbf{T}emporal \textbf{A}ccumulation (\modelname) score to depict the importance of each token. 
We take two factors into consideration, \textbf{temporal redundancy} and \textbf{semantic importance}. 
Our motivation is to discard tokens that have similar counterparts appearing earlier in the sequence while retaining only semantically significant tokens.
As an example, static backgrounds across all timestamps contain highly repetitive information that is unnecessary to be included. Therefore, keeping only a few representative background patches is sufficient for reasoning.
Specifically, we evaluate the temporal redundancy of a region by determining whether it is a novel or previously observed entity.
In practice, we aggregate repetitive information on a per-frame basis and assign each token a score of temporal repetition degree.
Nevertheless, there are cases where a repetitive region reveals a crucial action and should be retained. 
For example, if the sequence of tokens describes human-body motion, it is necessary to keep all the tokens for better understanding, even if there are only slight differences over time. 
Thus, we also take semantic importance into account during the pruning procedure.
To depict each token's semantic contribution to video recognition, we take the summation of the feature activation map and then integrate this summation with the score of temporal aggregation to enhance the awareness of semantics.
Based on the \modelname score, we progressively prune the tokens of the video Transformers three times.
The whole pruning process does not introduce any tuning parameter and directly accelerates the off-the-shelf video Transformers without the need to re-train. 

We apply our pruning strategy to two mainstream video Transformers, ViT~\cite{dosovitskiy2021an} and VideoSwin~\cite{liu2022video}, and evaluate $10$ off-the-shelf backbones on two action recognition benchmarks, Kinetics-400~\cite{kay2017kinetics} and Something-Something V2~\cite{goyal2017something}, to demonstrate the effectiveness of our method. As shown in Figure~\ref{fig:sota}, we achieve significant computation reduction with a negligible accuracy drop on Kinetics-400. For instance, using ViT-H as the backbone, by hierarchically pruning $57\%$ of the input tokens, \modelname reduces $49\%$ FLOPs while the accuracy drop is only $0.2\%$. Besides, with \modelname, FLOPs of VideoSwin-B are decreased by $38\%$ while maintaining $82.5\%$ accuracy with only a minimal drop of $0.2\%$. A similar trend can also be observed in the Something-Something V2 dataset.
Notably, we surpass STTS~\cite{wang2022efficient} by a $0.4\%$ accuracy gain with $40\%$ fewer FLOPs on Kinetics-400 and by a $0.5\%$ accuracy increase with $20\%$ fewer FLOPs on Something-Something V2 when using the same backbone.

% 
% compute-intensive resource-constrained.
% For example, the tokens of the same static background at all timestamps are surplus to be involved in the computation. Only one representative token kept is sufficient for reasoning. On the contrary, if the sequence of tokens represents dynamic motion, and needs to keep them all for better understanding.  

% \begin{figure}
%     \centering
%     \includegraphics[width=\linewidth]{Fig/teaser.pdf}
%     \caption{Motivation. Video data is redundant over time. New occurrence is sparse and rare. }
%     \label{fig:my_label}
% \end{figure}
% Avoiding the blindness on the class information,  we inject activation-based attention map $\mathcal{F}$ as indicated in Eqn.~\eqref{cam} to complement the accumulative temporal score.
% and pushes the efficiency limits,

\section{Related Work}
\paragraph{Video Transformers.} Designing Transformer-based architectures for vision tasks has emerged as a general trend in the computer vision community, as evidenced by several recent works~\cite{dosovitskiy2021an, liu2021swin, zhu2021deformable, patrick2021keeping, zheng2021rethinking, lu2021soft, wan2023seaformer}. With an unprecedented number of parameters and millions of training data, Transformers significantly outperform prior arts spanning a variety of tasks, not only in image but also in video understanding tasks. Various variants of self-attention have been introduced in prior works~\cite{bertasius2021space, neimark2021video, zhang2021vidtr, Arnab_2021_ICCV, bulat2021space, patrick2021keeping, fan2021multiscale, liu2022video, zha2021shifted, li2022mvitv2} to capture the spatiotemporal relationship. However, using pure patch-based Transformers incurs prohibitive costs on memory and computation when extracting global-range features from the whole video. 
To deal with it, 
Motionformer~\cite{patrick2021keeping} introduces trajectory attention that focuses on implicitly determined motion paths and optimizes the quadratic calculation via efficient decomposition. MeMViT~\cite{wu2022memvit} proposes caching `memory' of past frames and attending to the summarized prior context in an online manner. 
In this paper, we propose an orthogonal approach to make Transformers lighter by pruning the spatio-temporal tokens with high temporal redundancy.

\begin{figure*}
    \centering
    \includegraphics[width=\linewidth]{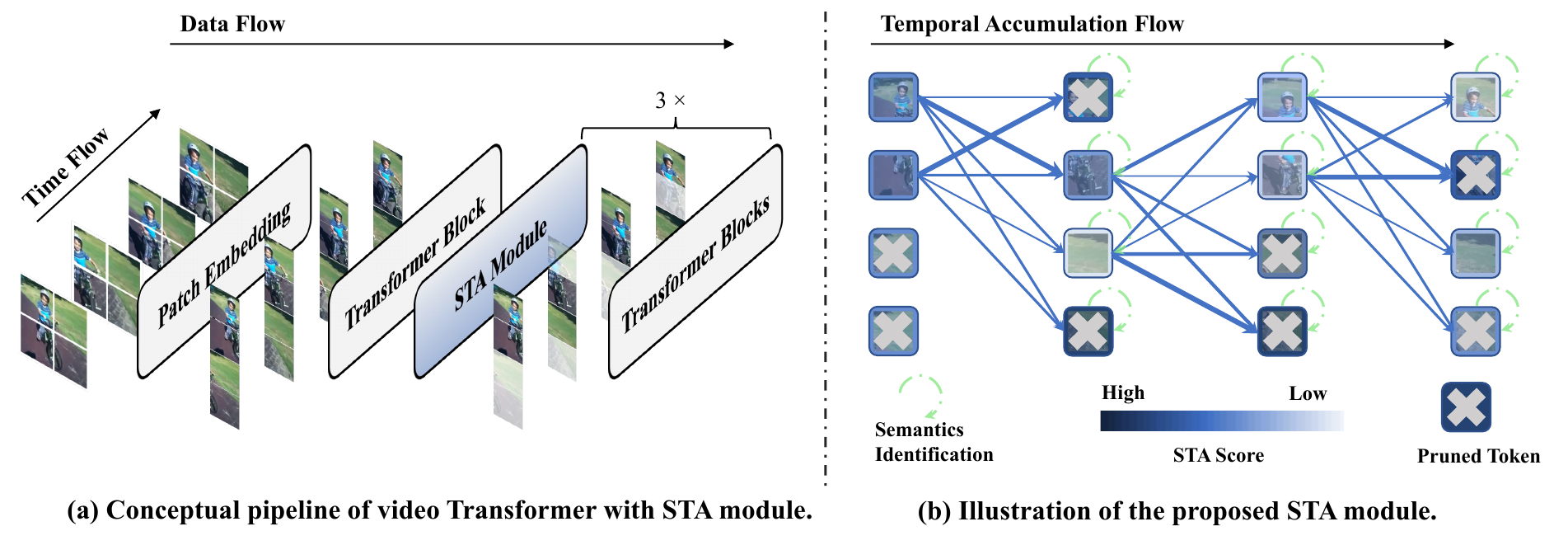}
    \caption{An overview of our \modelname token-pruning algorithm for video Transformers. (a) STA module is a simple plug-in and can be inserted at the beginning or end of the Transformer block. In practice, we conduct a three-stage progressive strategy to prune the token with \modelname. (b) Our semantic-aware temporal accumulation algorithm. The wider arrows connecting two adjacent frames represent higher weight to transport the STA score.}
    % \PS {``Time Flow'' in bold. This figure is better on Page 3.}
    \label{fig:overview}
\end{figure*}

\paragraph{Token pruning for Transformers.} Several works~\cite{ryoo2021tokenlearner, rao2021dynamicvit, wang2021not, liang2022evit, meng2022adavit, bolya2023token, li2023progressively} have focused on reducing the number of tokens involved in the calculation to accelerate image Transformer models. In specific, DynamicViT~\cite{rao2021dynamicvit} trains a lightweight decision module to rate the importance of each token and prune low-score tokens progressively. EViT~\cite{liang2022evit} preserves the attentive tokens guided by the class token attention and fuses inattentive ones without the help of any extra parameter. ToMe~\cite{bolya2023token} combines similar tokens to directly expedite off-the-shelf ViT without needing to train. 
While similar to ToMe~\cite{bolya2023token}, our method functions as a simple plug-in to enhance off-the-shelf video Transformers without requiring additional training. However, ToMe is an image-based pruning technique that does not model any temporal relations. In contrast, our method specifically devises a temporal aggregation mechanism tailored for video data.
% However, pruning tokens in video Transformers has received little attention.
Although STTS~\cite{wang2022efficient} trains a score network to choose pre-defined anchors from filtered frames within a video, the empirical speed-accuracy trade-off remains limited. This is because decoupled anchor-level selection still retains unwanted spatial redundancy.
On the contrary, we prune the video at the token level and eliminate the meaningless content by a large margin.

\paragraph{Efficient video recognition.} Due to the nature of the extra dimension, video understanding is computationally intensive. Thus, there have been attempts~\cite{zolfaghari2018eco, feichtenhofer2020x3d, tran2018closer, wu2019adaframe, korbar2019scsampler, wang2021adaptive, ding2022motion, Li_2020_CVPR, lee2018motion, lin2019tsm, jiang2019stm, qian2022static} to design lightweight modules that enjoy both high efficiency and high accuracy. 
% MFNet~\cite{lee2018motion} proposes a motion filter to efficiently extract spatiotemporal information. 
% R(2+1)D~\cite{tran2018closer} utilizes a mixture of 2D ConvNets and 3D ConvNets to jointly encode video data. 
ECO~\cite{zolfaghari2018eco} introduces a network architecture to sample a small subset of frames and learns the temporal context between these frames. Besides, AdaFocus~\cite{wang2021adaptive} improves the computational efficiency by adopting a light-weighted ConvNet to localize the most salient spatiotemporal regions. 
% Our work also belongs to efficient video recognition.    
While previous works have mainly concentrated on accelerating ConvNet models, efforts toward accelerating video Transformers have been relatively sparse and open to exploration. A recent approach~\cite{park2022k} devises a novel token-based sampling using k-centered search before feeding tokens into video Transformers. Although we also select semantically meaningful tokens for video Transformers, our dynamic model processes whole tokens at the early stage and prunes them based on model-dependent features.

\section{Approach}
The goal of this paper is to develop a principal token-pruning algorithm for video Transformers that achieves an optimal balance between cost, speed, and performance without requiring model re-training. 
We start by analyzing the video Transformers at hand and observe two interesting phenomena, detailed in Sec.~\ref{ins}.
\textit{First}, we find the high temporal redundancy when comparing the inter-frame similarity. \textit{Second}, the area that contributes to the final prediction usually takes up a small portion. 
Motivated by these two findings, we carefully develop two principles to prune the tokens with high temporal redundancy and retain the meaningful tokens. 
The overall framework is shown in Figure~\ref{fig:overview}(a). 
Our method is mainly built on the standard columnar Transformer~\cite{dosovitskiy2021an}, which we briefly go through in Sec.~\ref{vit}. Later, in Sec.~\ref{score}, we elaborate on the proposed metric to help prune unnecessary tokens. Finally,  Sec.~\ref{ins} discusses why \modelname works in video Transformer. 

\subsection{Revisit of Video Transformer}
\label{vit}
Video Transformers generally process video data as a 1D sequence of tokens and directly model the relationship between them. Initially, video Transformer linearly projects 3D data tubes into high-dimensional embeddings. 
Assuming the dimensions of video clips as $\{T, H, W\}$ and the size of 3D tubes as $\{t, h, w\}$, then the number of token embeddings is $n = n_t \times n_s = |\frac{T}{t}| \times |\frac{HW}{hw}|$. Additionally, positional embeddings are added to each token to break the permutation invariance.
After the patch embedding layer, an $n$-token sequence $\mathbf{X} \in \mathbb{R}^{n\times d}$ is passed into the self-attention layer, which computes a weighted sum of the values based on the affinity of other tokens. Mathematically, the self-attention is formulated as:
\begin{equation}
    \text{Attention}(\mathbf{Q},\mathbf{K},\mathbf{V}) = \text{softmax}(\frac{\mathbf{Q}\mathbf{K}^T}{\sqrt{d}})\mathbf{V},
\end{equation}
where $\mathbf{Q},\mathbf{K},\mathbf{V} = f_Q(\mathbf{X}), f_K(\mathbf{X}), f_V(\mathbf{X}) \in \mathbb{R}^d$ are typically linear transformations of $\mathbf{X}$. After spatial-temporal interaction, the tokens are sent into a feedforward network $f_{\text{FFN}}$, which consists of a three-layer MLP to exchange inter-channel information.

\subsection{Semantic-aware Temporal Accumulative Score}
\label{score}
Our intuitive criterion is to drop a token if similar tokens exist before it while reserving semantically meaningful tokens. 
To this end, our approach considers two factors when determining whether to retain or discard a token in the Transformer.
The first is its similarity to the other tokens along the temporal axis, and the second is its contribution to the class attribute. 
We discuss these two principles in order.
\paragraph{Temporal redundancy.} 
Intuitively, a token should be removed if similar tokens have already existed in previous frames.  
Therefore, we remove tokens frame-by-frame by comparing whether similar tokens have been retained. For simplicity, we reduce a constant number of tokens in each frame to ensure parallel computing. We introduce the accumulative temporal score $\mathbf{A}\in [0, 1]^{n_t \times n_s}$ to model the probability of dropping a token conditioned on the specific frame $t$. Specifically, we define:
\begin{equation}
    \mathbf{A}_{t, s} := \mathds{P}_{\text{drop}}(\mathbf{X}_{t,s}) \in [0, 1] \quad \mathrm{s.t.} \sum_{s=1}^{n_s}\mathbf{A}_{t, s} = 1,
\end{equation}
where tokens with higher temporal accumulative scores are more likely to be pruned because they carry a high degree of temporal redundancy.
Next, we eliminate $r$ tokens with the highest scores from $\mathbf{A}$ at $t$-th frame and transfer the remaining probability distribution $\mathbf{A}_{t}^{'} \in \mathbb{R}^{(n_s-r) \times 1}$ to the next frame via the transition probability $\mathds{P}_{\text{drop}}(\mathbf{X}_t|\mathbf{X}_{t-1}^{'})\in \mathbb{R}^{n_s \times (n_s-r)}$. By excluding the dropped tokens of the last frame, we effectively restart the repetition aggregation. This prevents the high scores from being concentrated on specific tokens and allows for the global identification of redundancy. 
Mathematically,
\begin{align}
\begin{split}
     &\mathbf{A}_{t+1} := \mathds{P}_{\text{drop}}(\mathbf{X}_{t+1}|\mathbf{X}_{t}^{'})  \mathbf{A}_{t}^{'}, \\
     &\mathds{P}_{\text{drop}}(\mathbf{X}_{t+1}|\mathbf{X}_t^{'})  := \text{softmax}(f(\mathbf{X}_{t+1}) f(\mathbf{X}_t^{'})^{T}), \\
     % &= \mathds{P}_{\text{drop}}(X_t|X_{t-1}) \mathds{P}_{\text{drop}}(X_{t-1}|X_{t-2}) A_{t-2} \\
     % &= \mathds{P}_{\text{drop}}(X_t|X_{t-1}) \cdots \mathds{P}_{\text{drop}}(X_{2}|X_{1}) A_{1}, \\  
\end{split}
\end{align}
where $f$ is the projection head to measure the similarity, and we construct transition probability by softmax-based affinity matrix. 
Note that we do not need to train a new projection head $f$ because the self-attention provides the necessary functionality, and the key function $f_K$ extracts most relevant knowledge for affinity estimation, as shown in Table~\ref{tab:sim_fun}. 
This formulation allows us to connect all temporally distinct tokens through a simple Markov chain and aggregate potential redundancy from the first frame to every subsequent frame. 
\begin{algorithm}[t]
\caption{Pseudocode of \textbf{\modelname} in a PyTorch-like style.}
\label{alg:code}
\algcomment{\fontsize{7.2pt}{0em}\selectfont \texttt{mm}: matrix multiplication.
%\vspace{-1.em}
}
\definecolor{codeblue}{rgb}{0.25,0.5,0.25}
\lstset{
  backgroundcolor=\color{white},
  basicstyle=\fontsize{7.2pt}{7.2pt}\ttfamily\selectfont,
  columns=fullflexible,
  breaklines=true,
  captionpos=b,
  commentstyle=\fontsize{7.2pt}{7.2pt}\color{codeblue},
  keywordstyle=\fontsize{7.2pt}{7.2pt},
%  frame=tb,
}
\begin{lstlisting}[language=python]
# x: token embedding, n_t x n_s x d
# I: image-based token pruning method
# r: drop number per frame
# sim: token-to-token affinity function

# min-max norm, Eqn.(4)
aam = norm(x.abs().sum(-1)) # size: (n_t, n_s, 1)

# token removal at 1-st frame 
x_0 = I(x[0]) # size: (n_s-r, d)
 
# initialization
x_list, x_old = [x_0], x_0
for t in range(1, n_t):
    # token-to-token affinity matrix
    A_t = sim(x[t], x_old) # size: (n_s, n_s-r)
    # accumulative temporal score
    s_acc = mm(A_t, s_acc) # size: (n_s, 1)
    # class-aware accumulative temporal score, Eqn.(5)
    s = s_acc * (1-aam[t]) 
    s = s.squeeze(dim=-1) # size: (n_s)

    # keep tokens with the minimal score
    i_t = s.topk(k=N-r, largest=False) # size: (n_s-r)
    x_old = x[t, i_t] # size: (n_s-r, d)
    x_list = x_list.append(x_old)
    
    # cut off the dropped tokens' score
    s_acc = s_acc[i_t] # size: (n_s-r, 1)
    # first-order norm
    s_acc = s_acc / s_acc.sum() 

return stack(x_list, dim=0) # size: (n_t, n_s-r, d)

\end{lstlisting}
\end{algorithm}
\paragraph{Semantic importance.}
Up until this point, our approach has focused on capturing temporally repetitive information. 
However, we have neglected the influence of semantic attributes. 
In other words, we have treated each token equally, regardless of its contribution to the semantics of the class. 
To integrate semantics importance with our approach, we use the activation-based attention map $\mathcal{F}$~\cite{zagoruyko2017paying}, which takes the feature matrix $\mathbf{X}\in\mathbb{R}^{n_t\times n_s\times d}$ as input and produces a score for each token in the matrix. Specifically, we define the semantic score for token $\mathbf{X}_{t,s}$ as:
\begin{equation}
    \mathcal{F}(\mathbf{X}_{t,s}) =  \sum_{i=1}^{d}|\mathbf{X}_{t,s,i}| \in \mathbb{R^{+}}.
    \label{cam}
\end{equation}
Intuitively, through the summation of absolute activation values over channel dimension, a high absolute activation suggests a significant contribution to next layers. Moreover, we apply STA on off-the-shelf Transformers supervised by semantic labels, where high activation areas tend to represent discriminative category information. Thus, activation-based attention maps could effectively capture the importance or relevance of the token to the overall semantics. 
We then use this score to re-weigh the temporal accumulative scores $\mathbf{A}$, giving tokens with high semantic contributions less weight in the pruning process. This ensures that tokens with high semantics are more likely to be retained, even if they have a high degree of temporal redundancy.

Finally, we compute the semantic-aware temporal accumulative score $\widetilde{\mathbf{A}}_{t,s}$ by integrating the semantic score $\mathcal{F}(\mathbf{X}_{t,s})$ with the accumulative temporal score $\mathbf{A}$, \textit{i.e.,}

\begin{equation}
    \widetilde{\mathbf{A}}_{t,s} = (1-\mathcal{F}(\mathbf{X}_{t,s})) \mathbf{A}_{t,s},
\end{equation}
where $\mathcal{F}(\mathbf{X}_{t,s})$ is min-max normalized to the range [0,1].
We utilize the semantic-aware accumulative temporal scores to guide token removal for all subsequent frames, except for the first frame.
Thus, we adopt an image-based token pruning method on the first frame to kick off our algorithm. 
Once tokens are discarded through our strategy, they are never employed in subsequent layers, thus accelerating the inference of the Transformer.

We summarize the pseudocode of \modelname in Alg.~\ref{alg:code}. 
The algorithm takes token embedding $\mathbf{X}\in \mathbb{R}^{n_t\times n_s\times d}$, an image-based token pruning method $I$, the number of tokens to drop per frame $r$, and a token-to-token affinity function as inputs. The algorithm calculates the \modelname score for each frame, selects the tokens with the minimal \modelname score, and retains them for the next frame. 
This process is repeated for all frames and returns the resulting token embedding matrix with the retained tokens $\mathbf{X'}\in \mathbb{R}^{n_t\times (n_s-r)\times d}$.

\paragraph{Summary of the superiority of STA.}
Compared to the previous token-pruning methods, our approach, \modelname, offers three significant merits when applied to video data:
\begin{itemize}
    \item \modelname fully considers the potential repetition of tokens along the temporal axis and eliminates the genuine redundancy with insignificant semantics. The temporal aggregation design makes the scoring mechanism more motion-aware and suitable for video data;
    \item \modelname works as a plug-in module without the introduction of additional parameters and it does not require the retraining of the video Transformer;

    \item \modelname achieves a complexity of $O\left(n_t n_s (n_s-r)\right)$, resulting in negligible additional FLOPs that only take up a small percentage of the total forward pass. Moreover, our algorithm allows for the bulk of computation to be done in parallel, making it friendly to modern GPU devices.
\end{itemize}

Overall, our approach is efficient and easily deployable, making it an ideal solution for pruning video Transformers.

%##################################################################################################
\subsection{Discussion}
\label{ins}
In this section, we present a two-part practical analysis to shed the light on the intuition behind \modelname. 
\paragraph{Does \modelname effectively reduce the temporal redundancy?}
To answer this question, we first define temporal redundancy as the frequency with which similar tokens appear at different timestamps.
We then assess this phenomenon in a video by probing the last frame, denoted as $\mathbf{X_{-1}}\in \mathbb{R}^{n_s\times d}$, and aggregating the cosine similarity between each token and the most similar tokens in previous frames. We term this aggregation as trajectory sum $S \in \mathbb{R}$.
Mathematically, 
\begin{equation}
    S=\frac{1}{n_s} \sum_{i=1}^{n_s} \sum_{t=1}^{n_t-1} \max_{j\in \{1,\cdots,n_s\}} \text{cos-sim}(\mathbf{X}_{-1}, \mathbf{X}_{t})_{ij},
\end{equation}
where $\text{cos-sim}(\mathbf{X}, \mathbf{Y})_{ij} = \frac{\mathbf{X}_i \cdot \mathbf{Y}_j}{|\mathbf{X}_i||\mathbf{Y}_j|}$.
A higher trajectory sum indicates greater temporal redundancy and lower diversity along the temporal axis. When all the frames are the same, the score would reach its theoretical maximum, which is $N_t-1$. 
Then, we compare our method with the standard Transformer and random-prune counterparts in terms of the proposed trajectory sum.  
Table~\ref{tab:trajectory} shows that the video Transformer exhibits heavy temporal redundancy and random pruning fails to alleviate them considerably. 
In contrast, \modelname achieves a far lower trajectory sum, indicating that it effectively eliminates temporal redundancy.

\begin{table}[]
\small
    \centering
    \begin{tabular}{l p{0.9cm}<{\centering}p{0.9cm}<{\centering}p{0.9cm}<{\centering}p{0.9cm}<{\centering}}
    \toprule
        Model & Small & Base & 
        Large & Huge\\\hline
        ViT & 5.10 & 5.38 &  5.07 &5.55\\
        Rand-ViT & 5.00 & 5.32 & 4.95 &5.46  \\
        STA-ViT  & 4.43 & 4.74 & 4.28 & 4.78\\
    \bottomrule
    \end{tabular}
    \caption{Trajectory sum for ViT family on the Kinetics-400 validation set. Compared to random pruning, STA-ViT decreases the temporal redundancy significantly.}
    \label{tab:trajectory}
\end{table}

\begin{figure}
    \centering
    \includegraphics[width=\linewidth]{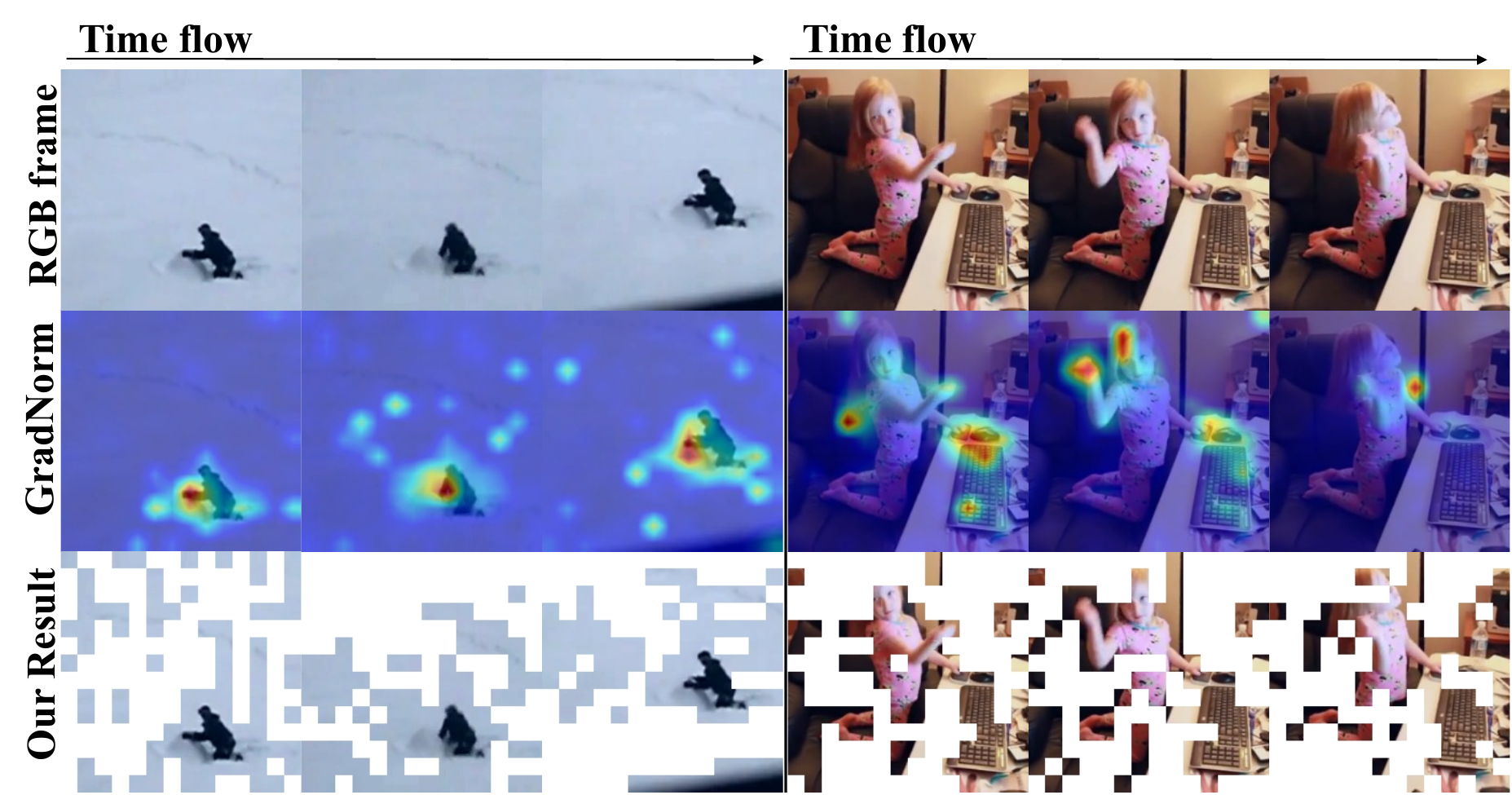}
    \caption{Gradient visualization for ViT-Large on the Kinetics-400 validation set. We stack original RGB frames, gradient norm heatmap, and our pruning result from top to bottom. Our pruning algorithm preserves the area of rich semantics well.}
     % (Left: making snowman; Right: pumping fist)
      
    \label{fig:grad_vis}
\end{figure}
\paragraph{Does \modelname retain semantics-rich tokens?}
To manifest it, we calculate the gradient norm for each token when back-propagating the training loss $\mathcal{L}$. 
% To fairly weigh the contribution in an end-to-end fashion, we aggregate the first-order gradient norm of $X^l$ at each layer $l$, \textit{i.e.,}
We then aggregated the first-order gradient norm of $\mathbf{X}^l$ at each layer $l$ to obtain the GradNorm, which reflects the contribution of each token to the final prediction.
Mathematically,
\begin{equation}
  \text{GradNorm}(\mathbf{X}) = \sum_{l=1}^{L}  \sum_{i=1}^{d}\left|\frac{\partial \mathcal{L}}{\partial \mathbf{X}^l_{\cdot,\cdot, i}}\right| \in \mathbb{R}^{n_t \times n_s},
\end{equation}
where $L$ is the total number of the Transformer block, {\em e.g.,} $L=24$ for ViT-Large. 
Figure~\ref{fig:grad_vis} shows the GradNorm distribution in the form of a heatmap. The heatmap reveals sparse patterns across the board, indicating that most tokens do not contribute significantly to the final prediction. Instead, the key regions responsible for the final prediction are usually motion-centric entities. This agrees with our intuition of highlighting semantically meaningful regions. 
% Figure~\ref{fig:grad_vis} visualizes the GradNorm in the format of the heatmap. The heatmap of GradNorm distribution shows sparse patterns across the board. Most tokens don't positively contribute to the final prediction. Rather, the key role responsible for final prediction only lies in minority regions, usually containing motion-centric entities. It coincides with our intuition of highlighting the semantics-rich region and discarding semantically meaningless regions. 
With the help of activation-based attention map $\mathcal{F}$ in Eqn.~\eqref{cam}, \modelname retains almost all areas of high-activation GradNorm as is evidenced in the last row of Figure~\ref{fig:grad_vis}. 

\begin{table*}[]
    \centering
    \begin{tabular}{p{2cm} p{3cm} p{2cm} p{2cm} p{2cm} p{2cm}} 
    \toprule
        \multirow{2}{*}{Base Model}  & \multirow{2}{*}{Metrics} & \multicolumn{4}{c}{Drop Number $r_1$}  \\
        \cmidrule(r){3-6}
        & & 0 & 32 & 48 & 64 \\\hline
        \multirow{3}{*}{ViT-S} & K400 Acc. (\%) & 78.8 & 78.8 \textcolor{blue}{(-0.0)}& 78.6 \textcolor{blue}{(-0.2)} & 78.1 \textcolor{blue}{(-0.7)} \\ 
        & SSV2 Acc. (\%) & 66.8  & 66.6 \textcolor{blue}{(-0.2)} & 66.4 \textcolor{blue}{(-0.4)} & 65.8 \textcolor{blue}{(-1.0)} \\ 
        & GFLOPs & 57 & 42 \textcolor{blue}{(-26\%)} & 35 \textcolor{blue}{(-39\%)}  & 29 \textcolor{blue}{(-49\%)}  \\\hline
        % & Throughput (clips/s) & 115 & 141 & 166 & \textbf{203} & 235 & 315 & 382\\
        \multirow{3}{*}{ViT-B} & K400 Acc. (\%) & 81.2 & 81.2 \textcolor{blue}{(-0.0)} & 81.1 \textcolor{blue}{(-0.1)} & 80.8 \textcolor{blue}{(-0.4)} \\ 
        & SSV2 Acc. (\%) & 70.6  & 70.4 \textcolor{blue}{(-0.2)}& 70.3 \textcolor{blue}{(-0.3)} & 69.9 \textcolor{blue}{(-0.7)} \\ 
        & GFLOPs& 180 & 136 \textcolor{blue}{(-24\%)}& 116 \textcolor{blue}{(-36\%)} & 96 \textcolor{blue}{(-47\%)}  \\\hline
        % & Throughput (clips/s) & 53 & 66 & 78 & \textbf{96} & 116 & 146 & 181\\
        \multirow{2}{*}{ViT-L} & K400 Acc. (\%) & 85.1 & 85.2 \textcolor{blue}{(+0.1)} & 85.1 \textcolor{blue}{(-0.0)} & 85.0 \textcolor{blue}{(-0.1)}\\ 
        & GFLOPs& 597 & 446 \textcolor{blue}{(-25\%)} & 376 \textcolor{blue}{(-37\%)} & 308 \textcolor{blue}{(-48\%)}\\\hline
        % & Throughput (clips/s) & 19 & 25 & 30 & 38 & \textbf{47} & 63 & 83\\
        \multirow{2}{*}{ViT-H} & K400 Acc. (\%) & 86.3 & 86.3 \textcolor{blue}{(-0.0)} & 86.2  \textcolor{blue}{(-0.1)} & 86.1 \textcolor{blue}{(-0.2)} \\ 
        & GFLOPs& 1192 & 890 \textcolor{blue}{(-25\%)} & 748 \textcolor{blue}{(-37\%)} & 611 \textcolor{blue}{(-49\%)} \\
        % & Throughput (clips/s) &  12 & 15 & 18 & 23 & \textbf{29} & 39 & 52\\
    \bottomrule
    \end{tabular}
    \caption{Main Results for \modelname-ViT family on Kinetics-400~\cite{kay2017kinetics} (K400) and Something-Something V2~\cite{goyal2017something} (SSV2). All input resolution is $16 \times 224^2$. }
    \label{result_vit}
\end{table*}

% \begin{figure*}
% \centering
% \begin{minipage}{.47\textwidth}
  % \centering
  % \includegraphics[width=\linewidth]{Fig/plot.pdf}
  % \caption{Trajectory Sum for ViT-Large on Kinetics-400 validation set. $r=64$ for both \modelname and random pruning methods.}
  % \label{fig}
% \end{minipage}%
% \hspace{5pt}
% \begin{minipage}{.47\textwidth}
%   \centering
%   \includegraphics[width=\linewidth]{Fig/curve.pdf}
%   \caption{Top-1 accuracy \texttt{vs.} throughput under two pruning methods with various prune numbers $r$. Rand-ViT-L means random pruning on ViT-L.}
%   \label{fig:sweet}
% \end{minipage}
% \end{figure*}

\begin{table*}
    \centering
    \begin{tabular}{p{2cm} p{3cm} p{2cm} p{2cm} p{2cm} p{2cm}}
    \toprule
        \multirow{2}{*}{Base Model}  & \multirow{2}{*}{Metrics} & \multicolumn{4}{c}{Drop Number $r_1$}  \\
        \cmidrule(r){3-6}
        & & 0 & 192 & 256 & 320  \\\hline
        \multirow{2}{*}{VideoSwin-T} & K400 Acc. (\%) & 78.8 & 78.7 \textcolor{blue}{(-0.1)} & 78.7 \textcolor{blue}{(-0.1)} & 78.6 \textcolor{blue}{(-0.2)}\\ 
        & GFLOPs & 88 & 68 \textcolor{blue}{(-23\%)} & 61 \textcolor{blue}{(-31\%)} & 54 \textcolor{blue}{(-39\%)}\\\hline
        \multirow{2}{*}{VideoSwin-S} & K400 Acc. (\%) & 80.5 & 80.3 \textcolor{blue}{(-0.2)} & 80.2 \textcolor{blue}{(-0.3)}& 80.1 \textcolor{blue}{(-0.4)}\\ 
        & GFLOPs & 166  & 121 \textcolor{blue}{(-27\%)} & 106 \textcolor{blue}{(-36\%)} & 91 \textcolor{blue}{(-45\%)} \\\hline
        % & Throughput (clips/s) & 115 & 141 & 166 & \textbf{203} & 235 & 315 & 382\\
        \multirow{4}{*}{VideoSwin-B} & K400 Acc. (\%) & 82.7 & 82.5 \textcolor{blue}{(-0.2)} & 82.5 \textcolor{blue}{(-0.2)}& 82.3 \textcolor{blue}{(-0.4)}\\
        & K400 GFLOPs & 282 & 202 \textcolor{blue}{(-28\%)} & 176 \textcolor{blue}{(-38\%)} & 149 \textcolor{blue}{(-47\%)} \\
        & SSV2 Acc. (\%) & 69.6 & 69.6 \textcolor{blue}{(-0.0)} & 69.5 \textcolor{blue}{(-0.1)} & 69.2 \textcolor{blue}{(-0.4)}\\ 
        & SSV2 GFLOPs & 321 & 241 \textcolor{blue}{(-25\%)} & 215 \textcolor{blue}{(-33\%)} & 188 \textcolor{blue}{(-41\%)} \\
        % & Throughput (clips/s) & 19 & 25 & 30 & 38 & \textbf{47} & 63 & 83\\
    \bottomrule
    \end{tabular} 
    \caption{Main Results for \modelname-VideoSwin family on Kinetics-400~\cite{kay2017kinetics} (K400) and Something-Something V2~\cite{goyal2017something} (SSV2). All input resolution is $32 \times 224^2$. VideoSwin-B employs varying window sizes for K400 and SSV2, leading to a discrepancy in FLOPs.}
    \label{result_swin}
\end{table*}
\section{Experiments}
\subsection{Experimental Setup} 
\paragraph{Datasets and backbones.} 

We evaluate our algorithm on two standard video action recognition datasets. Kinetics-400~\cite{kay2017kinetics} (K400) and Something-Something V2~\cite{goyal2017something} (SSV2). Kinetics-400 is a large-scale action recognition dataset sourced from YouTube, which consists of around 10-second clips with $400$ human action classes. The training and validation set has approximately $240$K and $20$K videos, respectively. Something-Something V2 is a motion-heavy benchmark with $174$ labels, where the object and background are shared across the different action categories. Around $170$K and $25$K videos exist in the training and validation set of SSV2, respectively. 
We implement our \modelname strategy with two mainstream Transformer-based backbones, namely ViT~\cite{dosovitskiy2021an} and VideoSwin~\cite{liu2022video}. For ViT, we sample $16$ frames with $224^2$ pixels as input and the size of 3D tube is $\{2, 16, 16\}$. Therefore, the number of tokens for all layers is $n=\frac{16}{2}\times\frac{224^2}{16^2}=8\times 14^2$. 
We load the open-sourced checkpoints from VideoMAE~\cite{tong2022videomae} due to its superior performance. For VideoSwin, the input resolution is $32\times 224^2$ and the size of 3D tube is $\{2, 4, 4\}$. The number of tokens at four hierarchical stages would be $\{16\times 56^2, 16\times 28^2, 16\times 14^2, 16\times 7^2\}$. 
Though our algorithm is not intended for window-based Transformers that require structural integrity, we still find a simple solution. We compensate for the discarded token with the nearest kept tokens when computing the window attention and speed up the rest of the process, such as the linear projection. In total, we test $10$ off-the-shelf backbones with different pre-trained weights on two benchmarks. 

% \PS{It is recommended to specify the video clip length settings for different datasets and backbones. This explains the discrepancy in GFLOPs between Table 3 and Table 2 across different datasets. It is also preferable to provide this information in the table captions.}
\paragraph{Implementation details.}
To progressively remove inattentive tokens, we apply our \modelname module three times. For ViT, we insert our \modelname module at the end of the $1^{st}$, $(1+L/3)^{th}$, $(1+2L/3)^{th}$ block, where $L$ is the total number of Transformer blocks. For VideoSwin, there are four hierarchical stages with varied resolutions and the \modelname module is located at the end of the first three stages. We denote different variants of \modelname as STA$^{r_1}$-Model, where $r_1$ is the number of spatial tokens dropped per frame at the first stage. We adopt a decreasing schedule, which reduces the dropped number by half at each stage. For instance, STA$^{64}$-ViT-L indicates pruning $\{8\times64,8\times32,8\times16\}$ tokens on ViT-Large. We weigh the computation via two metrics, FLOPs (floating-point operations) and throughput. FLOPs are reported with the help of fvcore library\footnote{\url{https://github.com/facebookresearch/fvcore}} and throughput (clips/s) is measured at a batch size of $32$ on a single 32G Tesla V100. Besides, we closely follow the inference metric for ViT and VideoSwin in~\cite{tong2022videomae, liu2022video}.
\begin{itemize}
    \item ViT~\cite{tong2022videomae}: To evaluate on K400, we sample 5 × 3 views by uniformly selecting 5 16-frame clips from a full-length video in the temporal dimension with the frame stride of 4. For each clip, we resize the shorter spatial side to 224 pixels and extract 3 crops of 224 × 224 resolution that cover the longer spatial axis. The final score is the average score computed over all views. For SSV2, we follow a similar procedure by sampling 2 clips × 3 crops and the frames stride is $2$.
    \item VideoSwin~\cite{liu2022video}: For K400, we extract 4 32-frame clips from each full-length video using a temporal stride of 2 and a spatial size of 224 × 224. Similarly, for SSV2, we extract 1 set of clips using a spatial size of 224 × 224 and 3 spatial crops, with a frame stride of 2. 
Besides, we prune VideoSwin three times as $\{r_1, 1.5r_1/4, 2r_1/16\}$. For instance, STA$^{256}$-VideoSwin-S indicates pruning $\{16\times256,16\times96,16\times32\}$ on VideoSwin-S.

\end{itemize}

% For more detail, please refer to our supplementary appendix.  

\subsection{Main Results}
% For example, DeiT-S trained with EViT with a keep rate of 0.7 increase the inference throughput by 50\% while maintaining the Top-1 accuracy reduction within 0.3% on ImageNet.
We conduct a thorough investigation of two off-the-shelf model families, ViT~\cite{dosovitskiy2021an} and VideoSwin~\cite{liu2022video}, on the Kinetics-400 and Something-Something V2 datasets. The results presented in Table~\ref{result_vit} demonstrate that our proposed method can significantly reduce the computational costs of ViT models by $25\%$ $\sim$ $49\%$, with negligible impacts on performance (-$0.2\%$ $\sim$ -$1.0\%$). It is worth noting our method shows a favorable trade-off between complexity and performance for larger models. For instance, our method reduces the FLOPs of ViT-Huge by half to just $611$ GFLOPs, with only a $0.2\%$ drop in accuracy.

To demonstrate the potential of our method to generalize well on various transformer backbones, we are conducting further experiments on VideoSwin~\cite{liu2022video}, a modern architecture that uses a window shuffling operation to interchange information. As VideoSwin is naturally unsuitable for unstructured tokens, we are filling the dropped locations during the window attention operation with the nearest tokens and then discarding the replicated tokens after the attention operation. Empirical results in Table~\ref{result_swin} indicate that the performance of VideoSwin holds until FLOPs fall by roughly $40\%$. This observation verifies that both columnar ViT and hierarchical VideoSwin have heavy and unnecessary computations that can be significantly optimized.

% We thoroughly investigate two off-the-shelf model families (ViT, VideoSwin) on K400 and SSV2. Table~\ref{result_vit} demonstrates that our proposed \modelname can help ViT reduce the computational costs by 25\% $\sim$ 49\% with the negligible influence of performance (-0.2\% $\sim$ -1.0\%). Note that we even do not tune the model. Besides, our method exhibits a favorable trade-off between complexity and performance at large models. For example, our method halves the FLOPs of ViT-Huge to a mere 611G but just drops 0.2\% accuracy. To demonstrate our method possesses the potential to generalize well on various transformer backbones, we conduct experiments on VideoSwin~\cite{liu2022video}, a modern architecture utilizing window shuffling operation to interchange the information. Considering VideoSwin is naturally unsuitable for unstructured tokens, we take the nearest tokens to fill the dropped location during the window attention operation. After the attention operation, we discard those replicate tokens. Empirical results in Table~\ref{result_swin} indicate that the performance of VideoSwin holds until FLOPs fall by roughly 40\%. This observation verifies that either columnar ViT or hierarchical VideoSwin owns heavy unnecessary computation that can be significantly optimized.

\begin{table}
    \centering
    \small
    \begin{tabular}{l l p{1cm}<{\centering}}
    \toprule 
       Model & GFLOPs$\times$views  &  Top-1\\\hline     
       TimeSformer-L~\cite{bertasius2021space}  & $8353\times1\times3$  & 80.7 \\
       Motionformer-L~\cite{patrick2021keeping}  & $1185\times10\times3$  &80.2\\
       ViViT~\cite{Arnab_2021_ICCV} &  $3981\times4\times3$ &  84.9  \\
       Swin-L~\cite{liu2022video} & $2107\times10\times5$& 84.9 \\
       MViTv2-L~\cite{li2022mvitv2} &  $2828\times5\times3$ & 86.1   \\
       ViT-H~\cite{tong2022videomae} & $1192 \times5\times3$ & 86.3 \\ 
       \rowcolor{mygray} STTS-VideoSwin-B~\cite{wang2022efficient} & $253 \times4\times3$ & 81.9 \\
       \rowcolor{mygray} ToMe-ViT-L~\cite{bolya2023token} & $281\times10\times1$  & 84.5  \\
       \rowcolor{mygray} \textbf{STA$^{320}$-VideoSwin-B (ours)} & $149 \times4\times3$ & 82.3\\
       \rowcolor{mygray} \textbf{STA$^{64}$-ViT-L (ours)} &  $308\times5\times3$  & 85.0 \\ %47  
       \rowcolor{mygray} \textbf{STA$^{64}$-ViT-H (ours)} &  $611\times5\times3$ & 86.1  \\ %29
    \bottomrule
    \end{tabular}
    \caption{Comparisons with the-state-of-the-arts method on Kinetics-400. We report the computational cost with a single view (temporal clip with spatial crop) × the number of views (FLOPs$\times$ view$_{\text{time}}$ $\times$ view$_{\text{space}}$). {\setlength{\fboxsep}{2pt}\colorbox{mygray}{Gray}} represents that this method leverages the dynamic token pooling to optimize existing backbones.}
    \label{tab:sota_k400}
\end{table}

\begin{table}[]
    \centering
    \small
    \begin{tabular}{l l p{1cm}<{\centering}}
    \toprule
       Model & GFLOPs$\times$views   &  Top-1\\\hline
       
       TimeSformer-L~\cite{bertasius2021space}  & $5549 \times1\times3$  & 62.4\\
       Motionformer-L~\cite{patrick2021keeping} & $1185 \times1\times3$ & 68.1 \\
       % UniFormer-B  & $96 \times 1 \times 3$ &70.2 &93.0\\
       MViTv2-B~\cite{li2022mvitv2}  & $225 \times1\times3$ & 70.5  \\
       VideoSwin-B~\cite{liu2022video} & $321 \times 1 \times 3$  & 69.6 \\
       ViT-B~\cite{tong2022videomae} & $180 \times2\times3$ & 70.6 \\
       \rowcolor{mygray}STTS-VideoSwin-B~\cite{wang2022efficient} & $237\times1\times3$ & 68.7 \\
       \rowcolor{mygray}\textbf{STA$^{320}$-VideoSwin-B (ours)}  & $188 \times1\times3$ &  69.2  \\
       \rowcolor{mygray}\textbf{STA$^{48}$-ViT-B (ours)}  & $116\times2\times3$ &  70.3  \\
    \bottomrule
    \end{tabular}
    \caption{Comparisons with the-state-of-the-arts method on Something-Something V2. We report the computational cost with a single view (temporal clip with spatial crop) × the number of views (FLOPs$\times$ view$_{\text{time}}$ $\times$ view$_{\text{space}}$). {\setlength{\fboxsep}{2pt}\colorbox{mygray}{Gray}} represents that this method leverages the dynamic token pooling to optimize existing backbones.}
    % We report the computational cost with a single view (temporal clip with spatial crop) × the number of views (FLOPs$\times$ view$_{\text{time}}$ $\times$ view$_{\text{space}}$). {\setlength{\fboxsep}{2pt}\colorbox{mygray}{Gray}} represents that this method leverages the dynamic token pooling to optimize existing backbones.
    \label{tab:sota_ssv2}
\end{table}

\paragraph{Comparison with the state of the art.}
% We have tabulated a comparison of our proposed method on Kinetics-400 in Table~\ref{tab:sota_k400}. Our model demonstrates superior performance in terms of both accuracy and computational cost. For instance, ViT-L equipped with our \modelname achieves the same accuracy as MViTv2-L~\cite{li2022mvitv2}, but with less than a quarter of the computational cost. Moreover, STTS~\cite{wang2022efficient} proposes a scorer network to perform dynamic token selection separately in space and time, requiring the scorer network to be trained in an end-to-end fashion. Our results surpass STTS by 0.4% accuracy using the same backbone VideoSwin-B, but with only 0.6$\times$ GFLOPs. This observation indicates that leveraging the model itself to weigh the importance of tokens is sufficient to reduce complexity.

% We also report our results on Something-Something V2 in Table~\ref{tab:sota_ssv2}. The superior performance of our proposed method verifies that our approach prunes inconsequential tokens via temporal cues, as it is known that understanding SSV2 heavily relies on temporal information. Specifically, ViT-B equipped with our \modelname surpasses most of the prior work with a considerably minor complexity of 116 GFLOPs. For VideoSwin, our strategy still outperforms STTS-VideoSwin by 0.5% accuracy with 80% of the computational cost.

Firstly, we tabulate a comparison of our proposed method on K400 in Table~\ref{tab:sota_k400}. Our model performs favorably in terms of both accuracy and computation cost. For example, ViT-L equipped with our \modelname achieves the same accuracy as MViTv2-L~\cite{li2022mvitv2} but with less than a quarter of the computational cost. 
Moreover, STTS~\cite{wang2022efficient} proposes a scorer network to conduct dynamic token selection separately in space and time, requiring to be trained in an end-to-end fashion. Our result surpasses STTS by $0.4 \%$ accuracy using the same backbone VideoSwin-B but with only $60\%$ GFLOPs. This result verifies that leveraging the model itself to weigh the redundancy of tokens is sufficient to reduce complexity. We also report the result on SSV2 in Table~\ref{tab:sota_ssv2}. The superior performance of our proposed method verifies that \modelname prunes the inconsequential tokens via temporal cues, as it is known that understanding SSV2 mainly relies on temporal information. Specifically, ViT-B equipped with our \modelname surpasses most of the prior arts with a considerably minor complexity of $116$ GFLOPs. For VideoSwin, our strategy outperforms STTS-VideoSwin by $0.5\%$ accuracy with $80\%$ of the computation cost.

\begin{figure}
    \centering
    \includegraphics[width=0.48\textwidth]{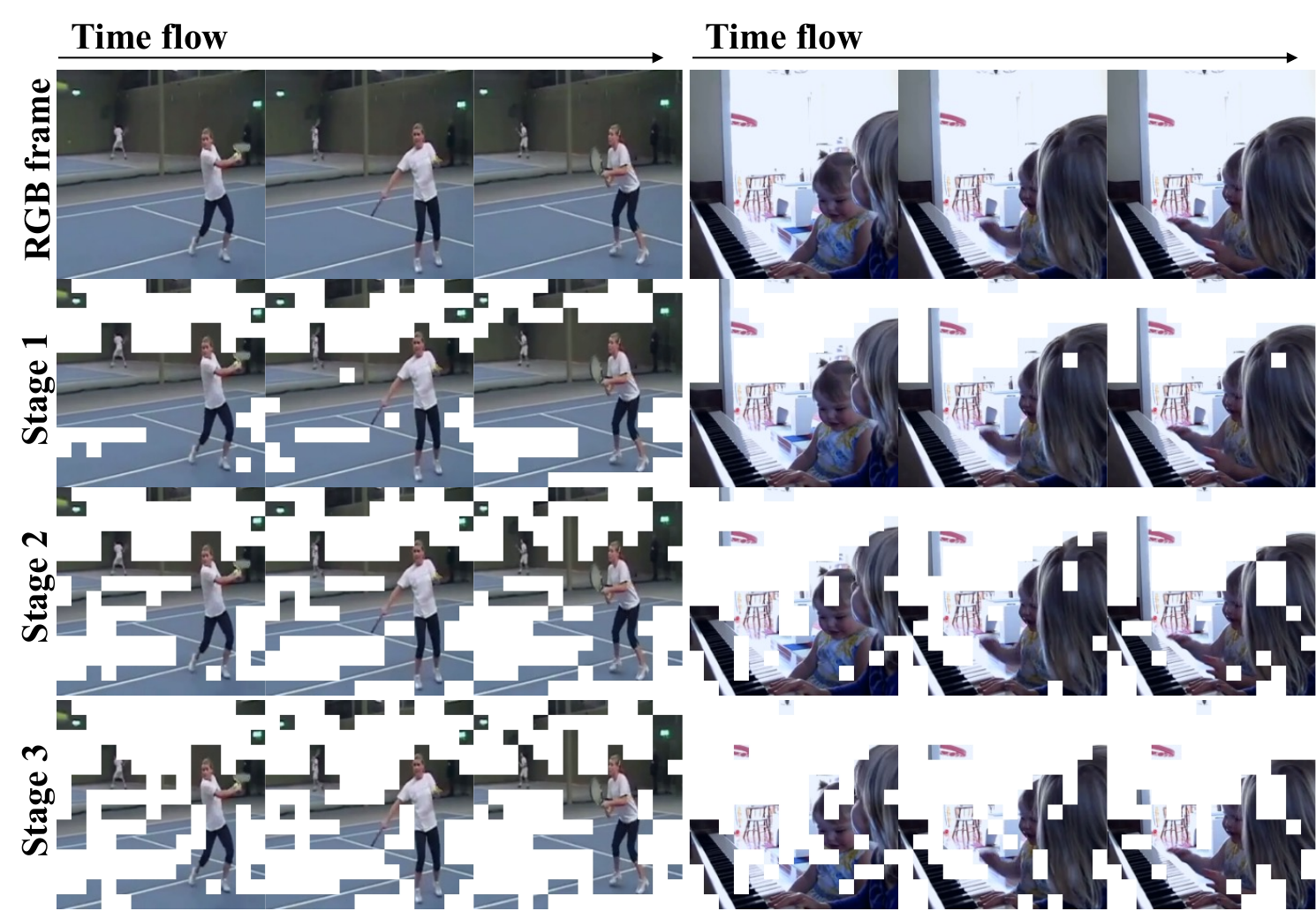}
    \caption{Visualization of the proposed \modelname strategy.  We masked out the discarded tokens with white boxes. STA not only retains informative tokens but also ensures diverse regions for improved video reasoning.}
    \label{fig:prune}
\end{figure}

\paragraph{Visualization of \modelname.}
Figure~\ref{fig:prune} shows image patches corresponding to kept tokens after three stages. The results align with our objective of resisting temporal redundancy and retaining informative tokens. In a tennis sequence, \modelname preserves the most meaningful patches, including a human at the far end of the court, and filters out dull backgrounds like the blue ground. The temporal aggregation design ensures that the kept tokens are not just the most salient ones but also a variety of regions, preserving diversity within videos for better reasoning.

%##################################################################################################
% overall table of all ablations
\begin{table*}[t]
\centering
\subfloat[ \normalsize
    Ablation on token removal methods at the first frame. 
    \label{tab:first-removal}
]{
    \centering
    \begin{minipage}{0.28\linewidth}{
        \begin{center}
        \begin{tabular}{l p{1cm}<{\centering} p{1cm}<{\centering}}
                \toprule
                    Method & Top-1 & Top-5  \\\hline
                    Random & 84.78 & 96.46\\
                    Grid & 84.85 & 96.50  \\
                    \rowcolor{mygray}ToMe & 84.96 & 96.50 \\%(216 + 432 + 864)
                \bottomrule \\
        \end{tabular}
    \end{center}}
    \end{minipage}
}
\hspace{0.1em}
\subfloat[ \normalsize
    Ablation on temporal accumulation order. 
    \label{tab:order}
]{
    \centering
    \begin{minipage}{0.28\linewidth}{
        \begin{center}
        \begin{tabular}{l p{1cm}<{\centering} p{1cm}<{\centering}}
                \toprule
                    Order &  Top-1 & Top-5 \\\hline
                    \rowcolor{mygray}F-B-F & 84.96 & 96.50 \\ 
                    B-F-B & 84.97 & 96.43 \\
                    F-F-F & 84.26 & 96.41 \\
                    B-B-B & 84.35 & 96.37 \\
                \bottomrule
        \end{tabular}
    \end{center}}
    \end{minipage}
}
\hspace{0.1em}
\subfloat[ \normalsize
    Ablation on different similarity function $f$.
    \label{tab:sim_fun}
]{
    \centering
    \begin{minipage}{0.28\linewidth}{
        \begin{center}
        \begin{tabular}{l p{1cm}<{\centering} p{1cm}<{\centering}}
         \toprule  
         Similarity &  Top-1 & Top-5\\\hline
         $f_Q$ &  84.83 & 96.52\\
         \rowcolor{mygray} $f_K$ & 84.96 & 96.50\\
         $f_V$ &  84.84 & 96.50\\
         $f_{\text{FFN}}$ & 84.94 & 96.50 \\\bottomrule
    \end{tabular}
    \end{center}}
    \end{minipage}
}
\\

\subfloat[ \normalsize
    Ablation on dropping schedule among three stages.
    \label{tab:schedule}
]{
    \centering
    \begin{minipage}{0.45\linewidth}{
        \begin{center}
\begin{tabular}{l p{1cm}<{\centering} p{1cm}<{\centering} p{1cm}<{\centering}}
        \toprule
            Schedule  & Top-1 & Top-5 & clips/s  \\\hline
            constant   & 84.68 & 96.36 & 47\\%(384 + 384 + 384)% 
            \rowcolor{mygray} decreasing    & 84.96 & 96.50  & 47\\%((512 + 256 + 128)
            increasing  & 77.68 & 93.72  & 44  \\%(216 + 432 + 864)
        \bottomrule
        \end{tabular}
    \end{center}}
    \end{minipage}
}
\hspace{0.5em}
\subfloat[ \normalsize
  Ablation on scoring mechanism. Top-1 is reported.
  \label{tab:score}
]{
    \centering
    \begin{minipage}{0.45\linewidth}{
        \begin{center}
        \begin{tabular}{l c c c}
        \toprule
            Score & ViT-S & ViT-B & ViT-L\\\hline
            % Random & 84.28 & 96.26 & 47.2 \\
            $1-\mathcal{F}(\mathbf{X}_{t,s})$ & 77.33 &  80.52 & 84.87  \\
            $\mathbf{A}_{t,s}$ &77.78 & 80.43 & 84.69 \\
            \rowcolor{mygray} $(1-\mathcal{F}(\mathbf{X}_{t,s}))\mathbf{A}_{t,s}$ & 78.12 & 80.82 & 84.96  \\
        \bottomrule
        \end{tabular}
    \end{center}}
    \end{minipage}
}
\\
%#################################################
\caption{Results of \modelname ablation experiments. F and B in (b) mean forward and backward order, respectively. The baseline ViT-L without \modelname obtains $85.05\%$ Top-1 and $96.55\%$ Top-5 accuracy on K400 at $19.5$ clips/s. {\setlength{\fboxsep}{2pt}\colorbox{mygray}{Gray}} is our default setting.}
\label{tab:ablations}
\end{table*}

\subsection{Ablation Study}
To find the optimal strategy, we conduct a series of ablation studies. We evaluate off-the-shelf ViT-Large on K400 by default and report accuracy, FLOPs, and throughput for reference unless otherwise stated.  

\paragraph{Token removal at the first frame.}  
To investigate how the first-frame removal affects performance, we conduct experiments on three candidates. (1) \textbf{Random Prune}: we randomly select $r$ tokens to discard. (2) \textbf{Grid Prune}: we split the first frame into $\sqrt{r}\times \sqrt{r}$ grids spatially and drop one random token per grid. (3) \textbf{ToMe Prune}: inspired by recent image pruning method ToMe~\cite{bolya2023token}, we get rid of the most similar tokens by the simple bipartite soft matching. The difference here is that we just drop tokens rather than `merge' tokens. Note that all three pruning ways are negligible in the terms of computation and lead to similar speed-up. It actually echoes the main insight of STA, leveraging temporal aggregation to reduce spatio-temporal redundancy. 
Even given a random initialization state, the sequential STA strategy could still decrease the total temporal redundancy and optimize the video Transformers effectively.
% Table~\ref{tab:first-removal} reveals that data-dependent removal is more suitable to initiate the \modelname algorithm though we still gain ideal performance using simple random pruning. 

\paragraph{Pruning schedule.}
We explore how to assign the number of dropping tokens among three stages. 
% As forwarding \modelname module once could exclude $n_t \times r$ tokens, we could drop $\{n_t \times r_1, n_t \times r_2, n_t \times r_3\}$ tokens sequentially. 
When maintaining a similar throughput, we devise three types of schedules: 
\begin{align*}
\begin{split}
    \text{Constant Schedule:} & \{8\times48,8\times48,8\times48\};  \\
    \text{Decreasing Schedule:} & \{8\times64,8\times32,8\times16\}; \\
    \text{Increasing Schedule:} & \{8\times27,8\times54,8\times108\}. \\
\end{split}    
\end{align*}
As shown in Table~\ref{tab:schedule}, the decreasing schedule owns the least accuracy reduction with similar throughput. It verifies that standard video Transformers process a great number of uninformative tokens that can be dropped at the beginning. 

\paragraph{Temporal accumulation order.}
Besides starting the accumulation flow from the beginning of the input video, we could also kick off at the ending frames. In Table~\ref{tab:order}, we empirically find alternating order at different dropping stages outperforms the consistent counterparts. We speculate that the same accumulation direction would amplify intrinsic propagation error but alternating the order counteracts it, leading to more reasonable pruning. 

\paragraph{Similarity function choice.} We ablate four similarity project heads $\{f_Q, f_V, f_K, f_{\text{FFN}}\}$. Table~\ref{tab:sim_fun} shows that key function $f_k$ captures the most correct affinity with minimal noise. 
The observation coincides with previous work~\cite{bolya2023token}.

\paragraph{Scoring mechanism.} To explore how accumulation score and semantic identification boost each other, we conduct the experiment with different scoring formulas. 
% Table~\ref{tab:score} demonstrates that considering both temporal redundancy and semantics helps discover the informative tokens. 
% The results on ViT-S show that temporal aggregation modeled by Markov Chain plays an important role in the pruning process. 
% While semantic importance functions effectively for ViT-B and ViT-L. 
Table~\ref{tab:score} demonstrates that considering both temporal redundancy and semantics helps in discovering informative tokens. The results on ViT-S show that temporal aggregation modeled by the Markov Chain plays an important role in the pruning process, while semantic importance functions effectively for ViT-B and ViT-L.

\begin{figure}
      \centering
      \includegraphics[width=0.95\linewidth]{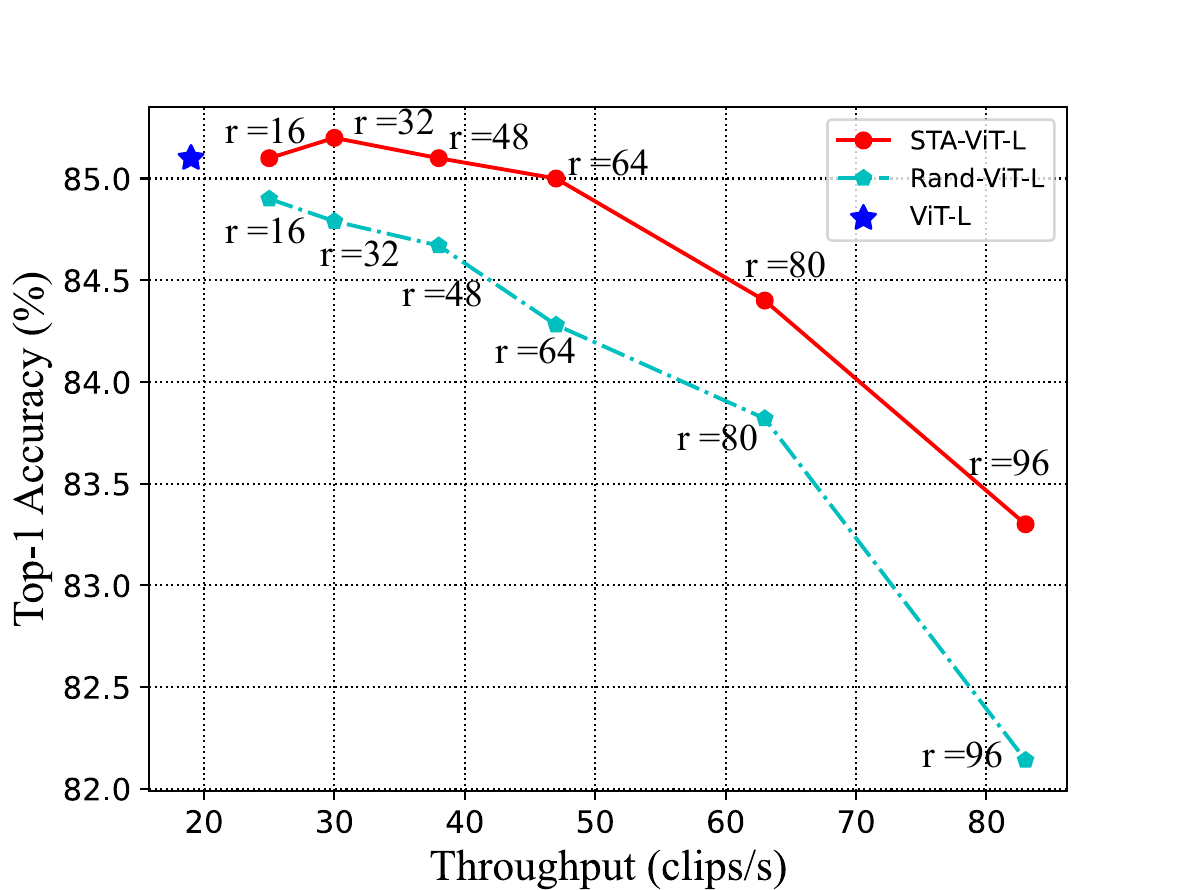}
      \caption{Top-1 accuracy and throughput under two pruning methods with various prune numbers $r$.}
      \label{fig:sweet}   
      \vspace{-1em}
\end{figure}

\paragraph{Performance \texttt{vs.} prune number $r$.}
To seek the sweet spot of our algorithm, we vary the prune number $r$ at the first stage ranging from [$16$, $96$] and evaluate the Top-1 accuracy. In addition, we compare our \modelname with the Random pruning baseline. As displayed in Figure.~\ref{fig:sweet}, \modelname behaves fairly robust to the token reduction and consistently surpasses the result of the random pruning. Specifically, $r=64$ doubles the throughput but just drops $0.1 \%$ accuracy. This confirms that our algorithm retains the semantics-rich tokens with the lowest redundancy. 
% We list overall results in Table~\ref{vit} and ~\ref{swin}.  

\section{Conclusion}
In conclusion, we propose a new token pruning strategy, Semantic-aware Temporal Accumulation (\modelname), for video Transformers that can significantly reduce computation overhead with a subtle accuracy drop. Specifically, we consider temporal redundancy and semantic importance when deciding to keep or drop the token.
Our approach does not introduce any parameter and can directly accelerate the off-the-shelf video Transformers without training.   
The extensive experiments demonstrate that our method empowers video Transformers to obtain a competitive speed-accuracy trade-off compared to the prior arts. 

\section*{Acknowledgement}
This work was supported in part by the National Natural Science Foundation of China under Grant 62250055, Grant 61932022, Grant 62120106007, and in part by the Program of Shanghai Science and Technology Innovation Project under Grant 20511100100. 

\clearpage
{\small
\bibliographystyle{ieee_fullname}
\bibliography{egbib}
}

\clearpage
\appendix
\section{More Experimental Results}
\paragraph{Training details.}
We also train the ViT-B with our proposed \modelname. We adopt dense sampling~\cite{wang2018non, feichtenhofer2019slowfast} on K400. We sample $16$ consecutive frames with the stride of $4$. The resolution is $224\times 224$. 
We perform RandAug augmentation (9, 0.5)~\cite{cubuk2020randaugment}, label smoothing (0.1)~\cite{szegedy2016rethinking},
mixup (0.8)~\cite{zhang2018mixup},
cutmix (1.0)~\cite{yun2019cutmix}, and random horizontal flip (0.5). In addition, we
adopt the repeated augmentation~\cite{hoffer2020augment}.
With DeepSpeed~\footnote{\url{https://github.com/microsoft/DeepSpeed}}, We use the linearly scale scheme to ensure effective parameter updates across different batch sizes during training, {\em i.e.,} $lr$ = base learning rate $\times$ batch size / $256$.
Specifically, we use the AdamW optimizer with a base learning rate of $1$e$-3$ and weight decay of $0.05$.
Beside, using a cosine decay learning rate scheduler and 5 epochs of linear warm-up, we finetune the model for 100 epochs with a total batch size of 128 on 4 nodes of 8 Tesla V100 GPUs. 
\paragraph{Training results.}
Besides speeding up the inference of off-the-shelf backbones, our algorithm also has the potential to expedite training. We report the training hours for ViT-Base in Table~\ref{tab:train}. \modelname cuts the training time in half. Without modifying the training recipe, the trained model only drops 0.6 \% in Top-1 accuracy. We believe that \modelname would be more effective to maintain the performance when training deeper backbones. We leave it as the future work.
\paragraph{Number of views.}
To analyze the impact of the number of test clips on our method, we conduct an experiment by varying the number of clips and comparing the results with the baseline ViT-L model. 
In Table.~\ref{num_clips}, we show that the relative performance drop remains constant at approximately 0.1\% regardless of the number of views, when the drop number is set to $r_1=64$. Furthermore, when using a lower value of $r_1=48$, there is no significant decrease in performance compared to the baseline.

\paragraph{Number of STA blocks and insert location}
We devise two extra ablation studies shown in Table~\ref{tab:sta}. Our experiments demonstrate that incorporating 3 progressive blocks at the very first beginning achieves an optimal trade-off. This approach allows for preferable computation while delivering maximal performance. 

\begin{table}[htbp]
    \centering
    \begin{tabular}{lccc}
    \toprule
        Model & clips/s & Training time & Top-1  \\\hline
        ViT-B & 53 &  28 hrs  & 81.2 \\
        STA$^{48}$-ViT-B & 96 & 15 hrs & 80.6 \\
    \bottomrule
    \end{tabular}
    \caption{Comparison on training time on Kinetics-400. We measure training time on 4 nodes of 8 V100.}
    \label{tab:train}
\end{table}

\begin{table}[]
    \centering
    \begin{tabular}{c c c c c}
    \toprule
        \multirow{2}{*}{Views} & \multicolumn{4}{c}{Drop Number $r_1$}  \\
        \cmidrule(r){2-5}
        & 0 & 48 & 64 & 80\\\hline
        2 $\times$ 3  & 83.36 &  83.21 &  83.09 &  82.56\\
        4 $\times$ 3 & 85.10 & 85.00 & 84.85 & 84.35 \\
        6 $\times$ 3& 85.05 &  85.07 & 84.84 & 84.59  \\
        8 $\times$ 3& 84.91 &  84.93 & 84.80 &  84.43 \\
        16 $\times$ 3& 84.91 & 84.97 & 84.89 & 84.48\\
    \bottomrule
    \end{tabular}
    \caption{Ablation on the temporal views of test clips.}
    \label{num_clips}
\end{table}

\begin{table}[]
    \centering
    \small
    \begin{tabular}{cll|lll}
        \toprule
         \# of STA & GFLOPs & Top-1 & Location & GFLOPs & Top-1  \\\hline
         2 & 302 & 84.5  & 1,9,17 & 308 & 85.0 \\
         3 & 308 & 85.0 & 3,11,19 & 339 & 85.0 \\
         4 & 305  & 84.8 & 5,13,21 & 370 & 85.1 \\
         \bottomrule
    \end{tabular}
    \caption{Ablation on the number of STA blocks and insert location.}
    \label{tab:sta}    
\end{table}
\section{More Visualization}
\begin{figure*}
    \centering
    \subfloat{
    \includegraphics[width=\linewidth]{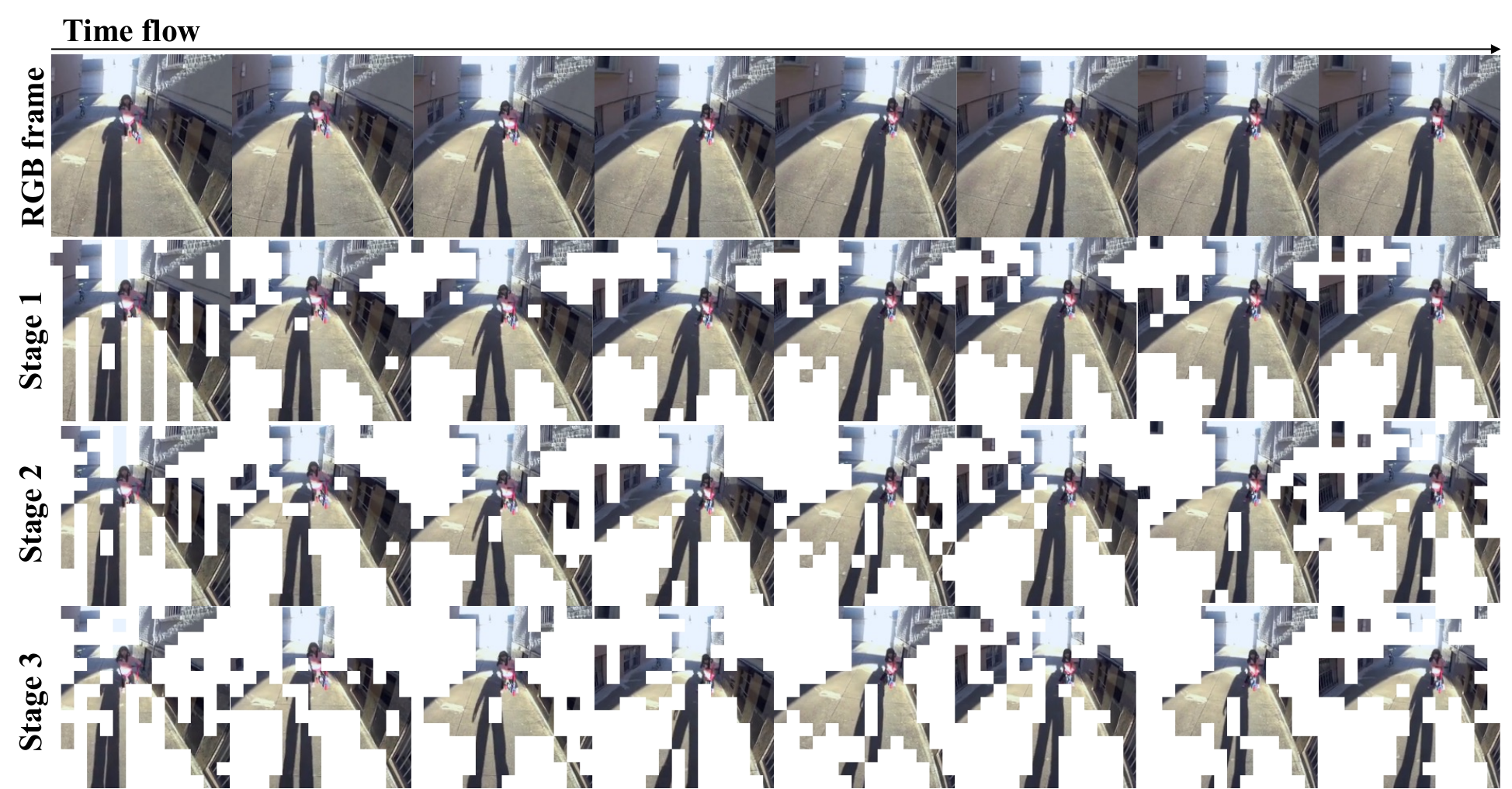}
    }\\
    \subfloat{
    \includegraphics[width=\linewidth]{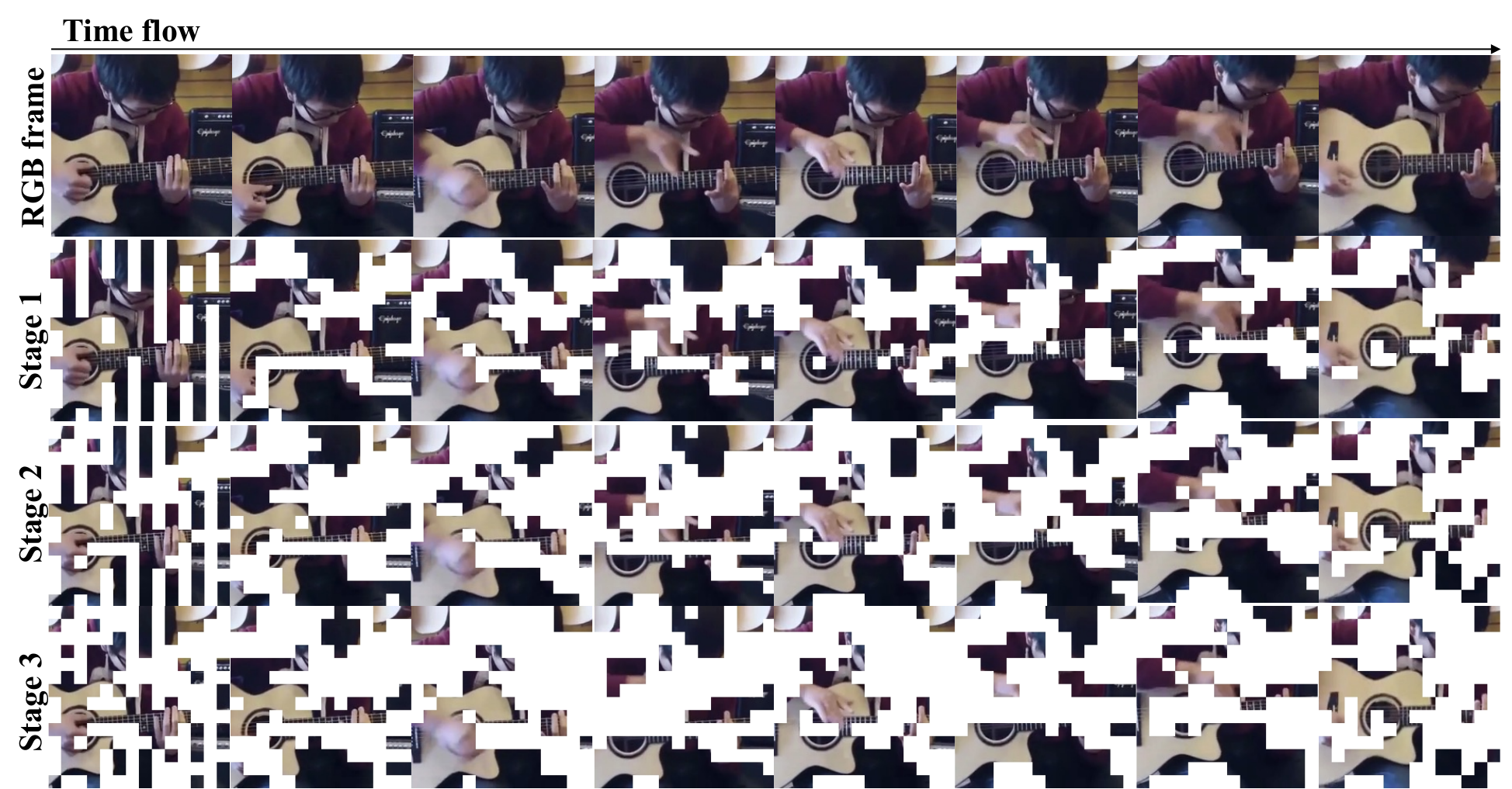}
    }
    \caption{Visualization of our STA strategy on K400.}
    \label{fig:k400}
\end{figure*}

\begin{figure*}
    \centering
    \subfloat{
    \includegraphics[width=\linewidth]{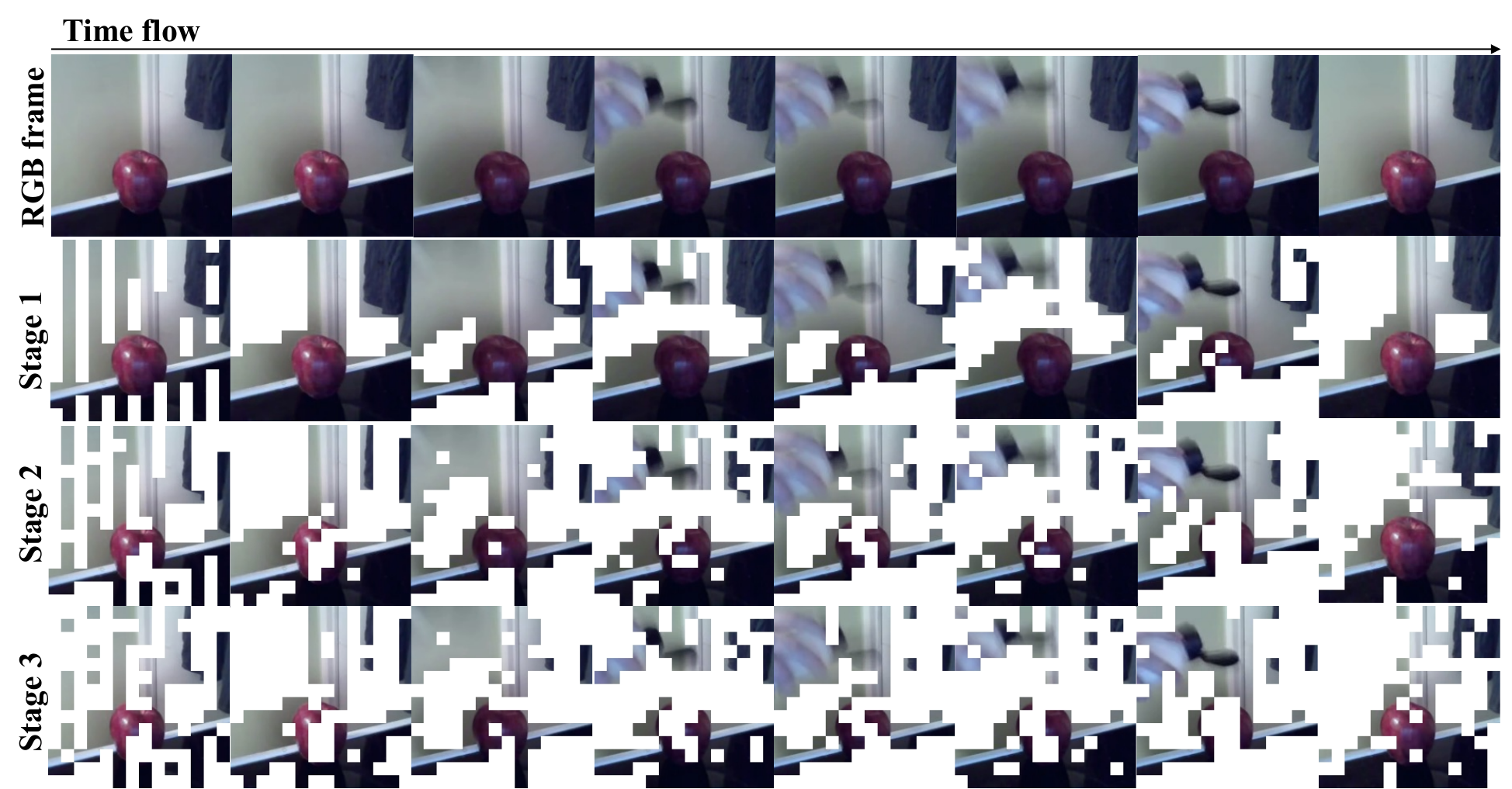}
    }\\
    \subfloat{
    \includegraphics[width=\linewidth]{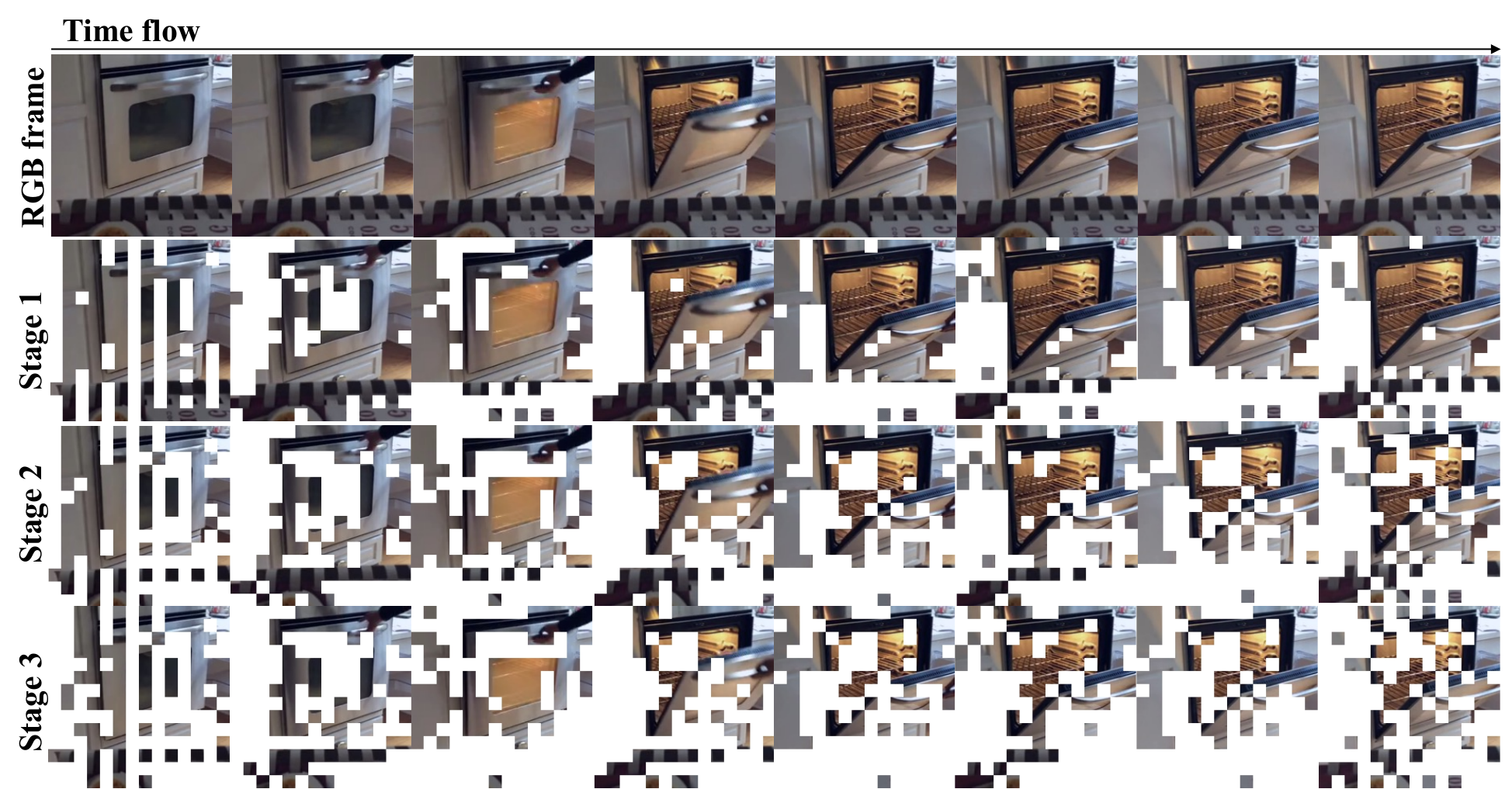}
    }
    \caption{Visualization of our STA strategy on SSV2.}
    \label{fig:ssv2}
\end{figure*}

We provide more visualization for our \modelname on K400 in Figure~\ref{fig:k400} and SSV2 in Figure~\ref{fig:ssv2}, which display image patches that correspond to the tokens retained after three stages of pruning. 
We observe that the pruning results align well with our objective of preserving detail-rich tokens and resisting temporal redundancy. 
Specifically, upon examining the guitar-playing sequence in Figure~\ref{fig:k400}, \modelname accurately preserves two partially visible guitars on the wall.
Additionally, the dropped tokens shown in Figure~\ref{fig:ssv2} at different timestamps are distributed unevenly, preserving the diversity of the video content.

\begin{figure*}
    \centering
    \subfloat{
    \includegraphics[width=\linewidth]{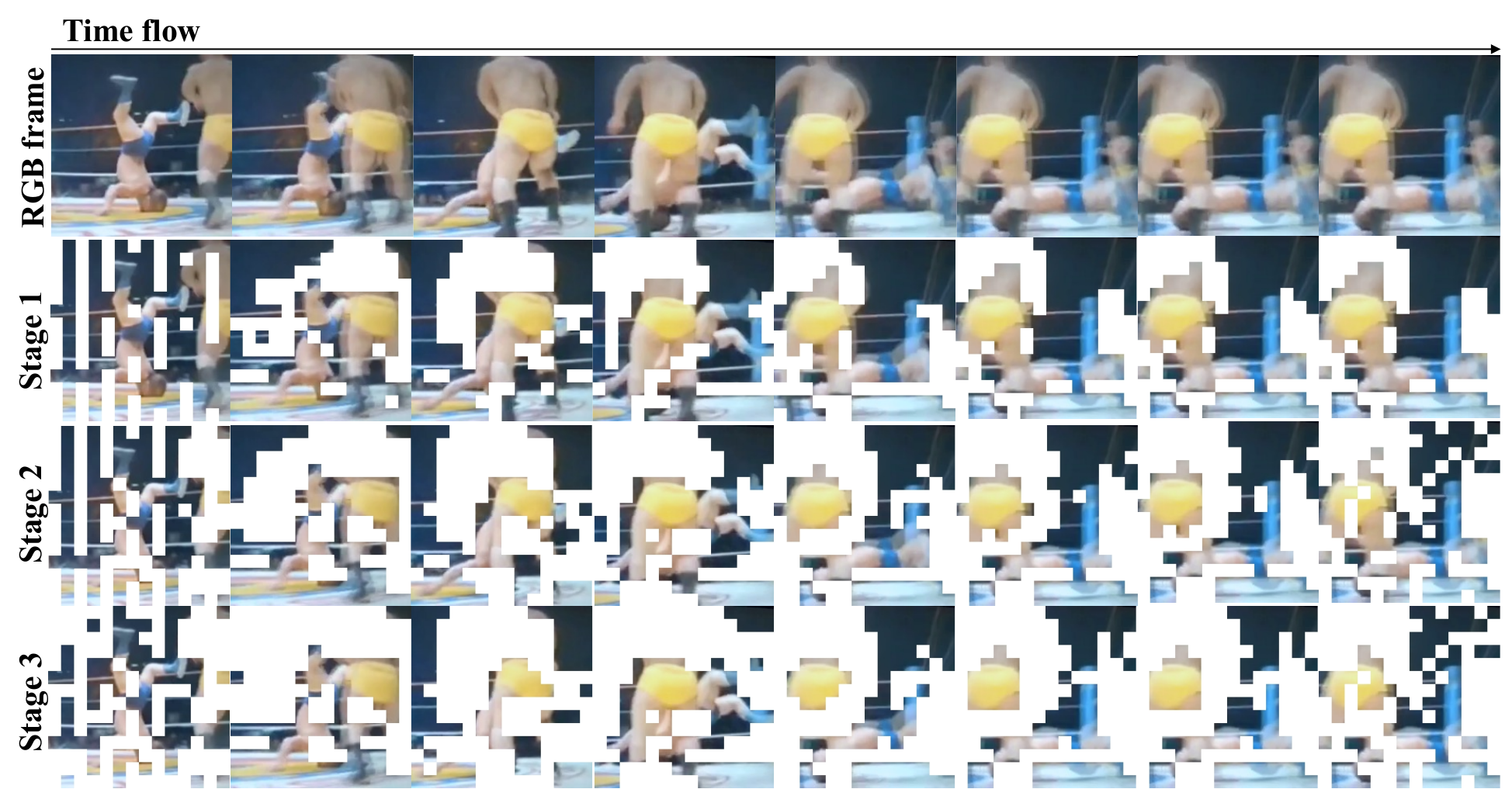}
    }\\
    \subfloat{\includegraphics[width=\linewidth]{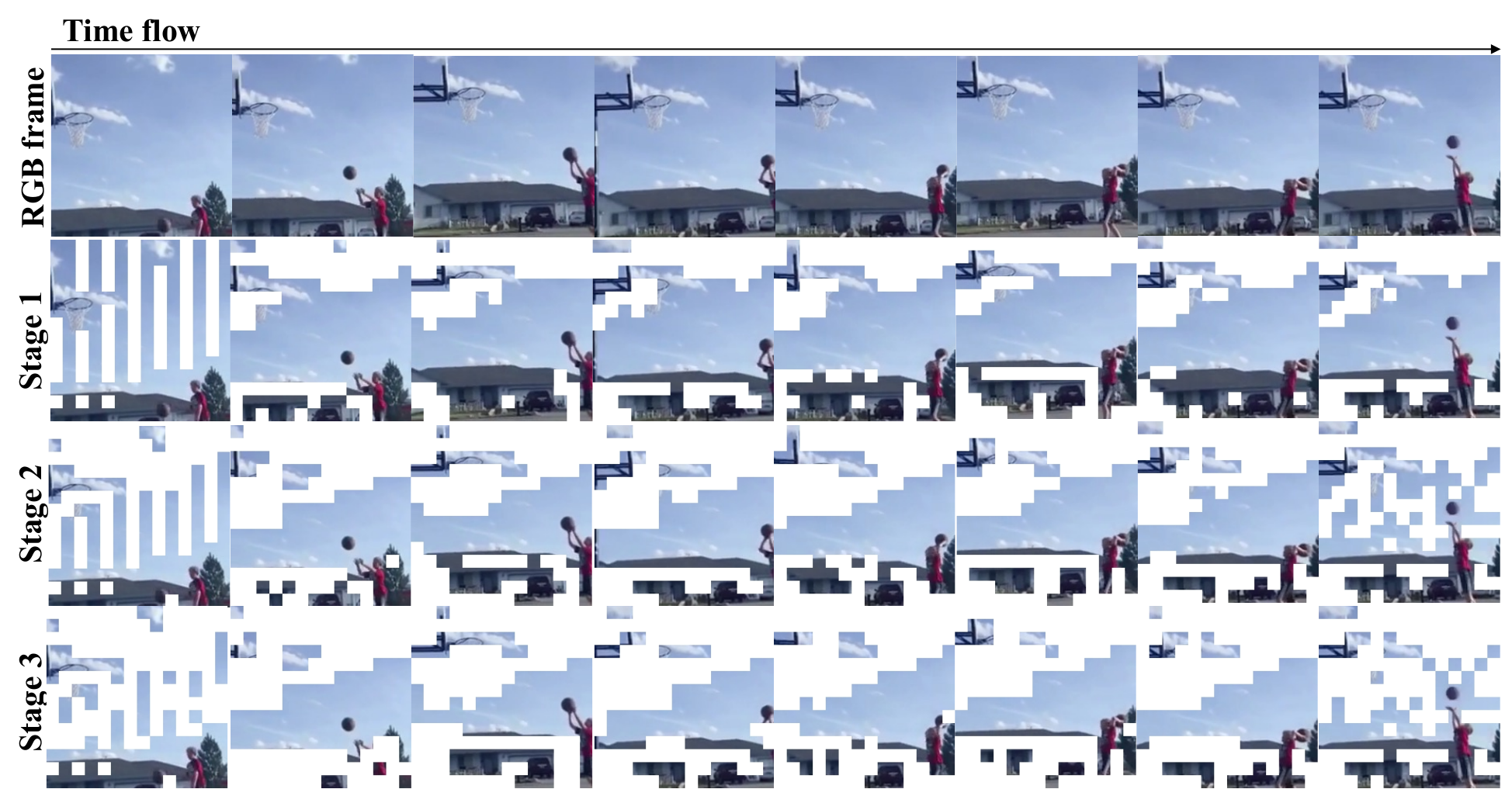}}
\end{figure*}
\newpage
\newpage
\begin{figure*}
    \centering
    \subfloat{
    \includegraphics[width=\linewidth]{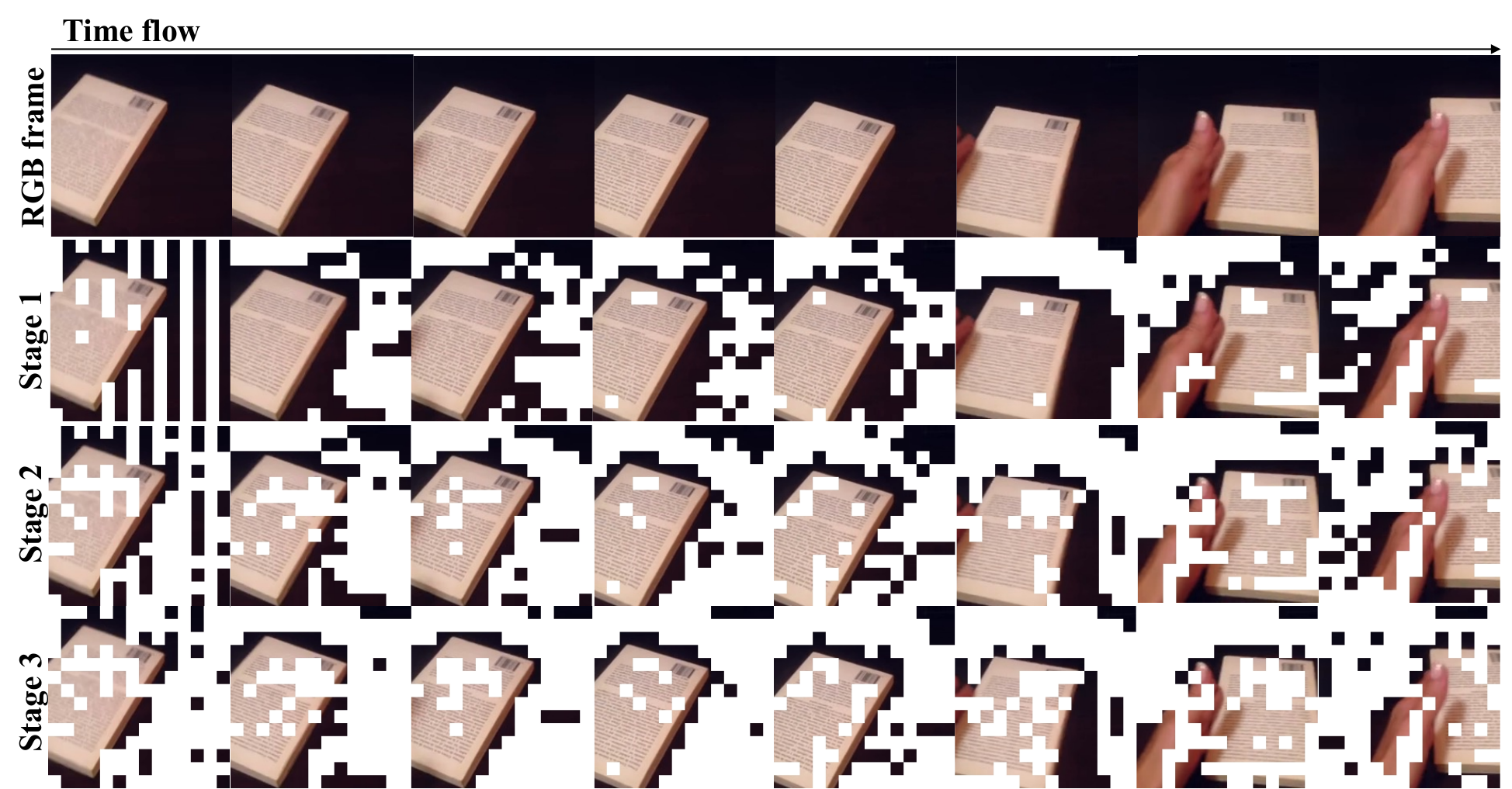}
    }\\
    \subfloat{\includegraphics[width=\linewidth]{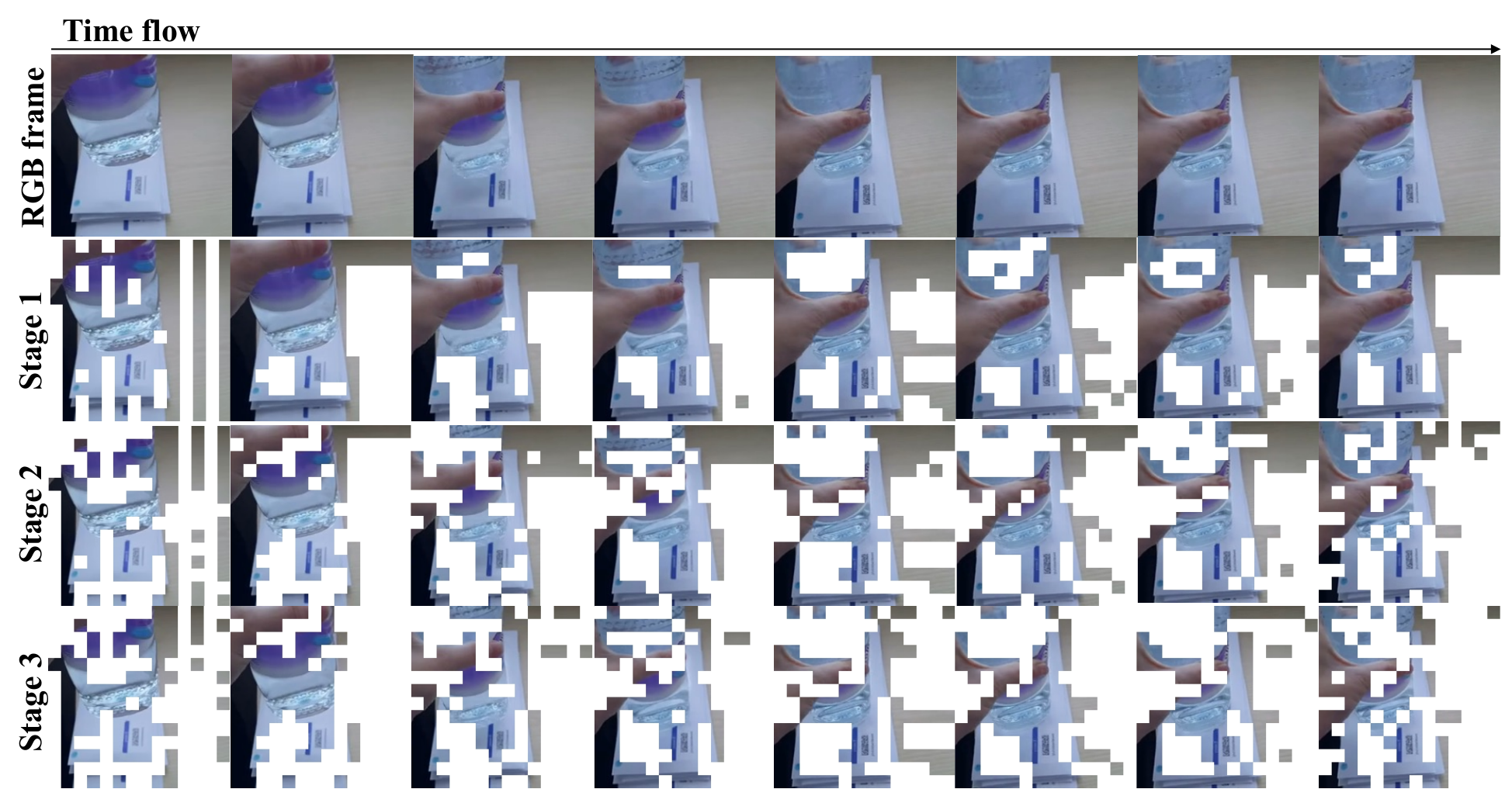}}
\end{figure*}
\newpage

\end{document}